\documentclass[10pt,twocolumn,letterpaper]{article}
\usepackage[pagenumbers]{cvpr}
\definecolor{cvprblue}{rgb}{0.21,0.49,0.74}
\usepackage[pagebackref,breaklinks,colorlinks,allcolors=cvprblue]{hyperref}
\usepackage{multirow}
\usepackage{makecell}
\usepackage{microtype}
\usepackage{titletoc}
\usepackage{tocloft}

\usepackage[many,listingsutf8]{tcolorbox}
\usepackage[frozencache,cachedir=minted-cache]{minted} 
\usepackage{marvosym}
\usepackage{amsmath}

\title{OmniMoGen: Unifying Human Motion Generation via Learning from Interleaved Text-Motion Instructions}

\author{
\parbox{\textwidth}{
\centering
\textbf{Wendong Bu}$^{1*}$\quad
\textbf{Kaihang Pan}$^{1*}$\quad
\textbf{Yuze Lin}$^{1*}$\quad
\textbf{Jiacheng Li}$^{2}$\quad
\textbf{Kai Shen}$^{1}$ \\
\textbf{Wenqiao Zhang}$^{1}$\quad
\textbf{Juncheng Li}$^{1\text{\Letter}}$\quad
\textbf{Jun Xiao}$^{1}$\quad
\textbf{Siliang Tang}$^{1}$ \\
\vspace{0.5em}
$^{1}$Zhejiang University\quad
$^{2}$HiThink Research \\
\texttt{\{wendongbu, junchengli\}@zju.edu.cn}
}
}

\begin{document}
\addtocontents{toc}{\protect\setcounter{tocdepth}{-1}}

\twocolumn[{%
\renewcommand\twocolumn[1][]{#1}%
\maketitle

\includegraphics[width=\linewidth]{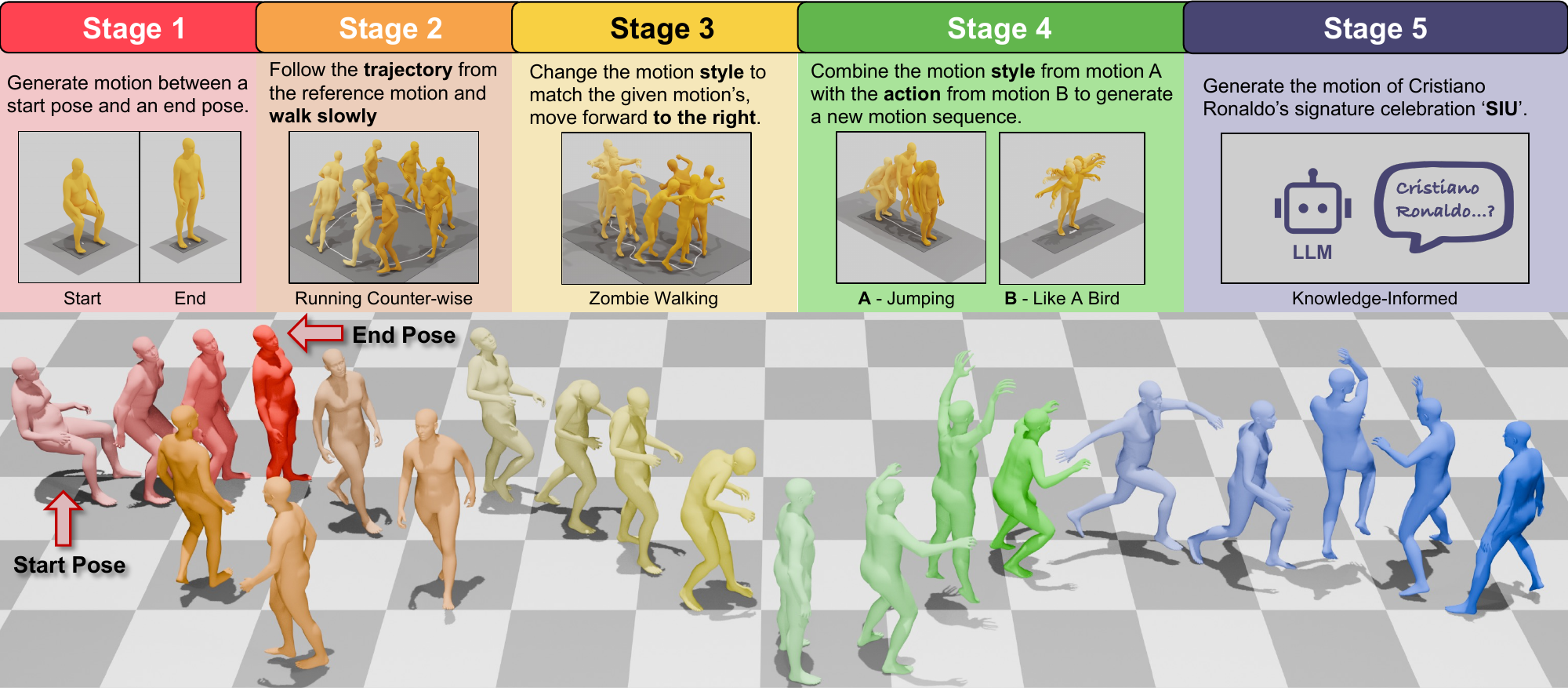}
\captionof{figure}{Similar to ChatGPT in NLP, OmniMoGen unifies all motion generation tasks in a unified architecture, such as text-to-motion, style editing, trajectory editing, inpainting, in-betweening, compositional editing, self-reflective generation, and knowledge-informed generation. OmniMoGen enables seamless and flexible motion generation across diverse objectives by merely adjusting the interleaved text-motion instructions. \vspace{1em}}
\label{fig:intro}
}]

{
  \makeatletter
  \renewcommand\@makefnmark{} 
  \long\def\@makefntext#1{\parindent 1em\noindent #1}
  \makeatother
  \footnotetext{\quad~*~~~~Equal Contribution.}
  \footnotetext{\quad\Letter~~~Juncheng Li is the Corresponding Author.}
}

\begin{abstract}
Large language models (LLMs) have unified diverse linguistic tasks within a single framework, yet such unification remains unexplored in human motion generation. Existing methods are confined to isolated tasks, limiting flexibility for free-form and omni-objective generation. To address this, we propose OmniMoGen, a unified framework that enables versatile motion generation through interleaved text-motion instructions. Built upon a concise RVQ-VAE and transformer architecture, OmniMoGen supports end-to-end instruction-driven motion generation. We construct X2Mo, a large-scale dataset of over 137K interleaved text-motion instructions, and introduce AnyContext, a benchmark for evaluating interleaved motion generation. Experiments show that OmniMoGen achieves state-of-the-art performance on text-to-motion, motion editing, and AnyContext, exhibiting emerging capabilities such as compositional editing, self-reflective generation, and knowledge-informed generation. These results mark a step toward the next intelligent motion generation. Project Page: \url{https://OmniMoGen.github.io/}.
\end{abstract}    
\section{Introduction}
\label{sec:intro}

Human motion generation plays a crucial role in animation, embodied intelligence, gaming, and virtual reality, including various tasks such as text-to-motion, motion editing, inpainting, and in-betweening.
Existing motion generation methods typically treat these tasks as independent problems with distinct architectures and training objectives.
For example, mainstream masked~\cite{guo2023momaskgenerativemaskedmodeling, pinyoanuntapong2024mmmgenerativemaskedmotion,li2025lamplanguagemotionpretrainingmotion} and diffusion-based~\cite{tevet2022humanmotiondiffusionmodel,athanasiou2024motionfixtextdriven3dhuman,jiang2025dynamicmotionblendingversatile} generative methods are designed to solve specific tasks such as text-to-motion, style transfer, and body inpainting.
This makes them struggle to follow arbitrary user instructions within a single architecture for omni-generation, as shown in Figure~\ref{fig:intro}.

Such a lack of unification in motion generation brings two major issues: 
\textbf{First, it limits the free-form motion generation.} 
Common isolated generation tasks, such as text-to-motion or body-part editing, rely on task-specific input forms, making them inherently limited to a narrow generation objective. 
However, real-world scenarios call for the next intelligent motion generation, where users can naturally generate various human motions through free-form instructions.
\textbf{Second, it hinders a unified model architecture.}
When generation tasks are not unified, introducing a new task to an existing model typically requires designing additional modules. 
This makes it difficult to unify diverse generation tasks into a single model.

Considering the two issues mentioned above and inspired by the success of Large Language Models~(LLMs) in unifying natural language processing~(NLP) tasks, we pose the following question: \textbf{Is it possible for a single model to handle omni-generation tasks about human motion end-to-end via user instructions, similar to the way \mbox{ChatGPT} handles linguistic tasks?} 
We envision a future where motion generation becomes intuitive and flexible, allowing users to accomplish any task directly through natural instructions. 
Guided by this vision, we introduce \mbox{\textbf{OmniMoGen}}, a unified framework for versatile motion generation. 
As shown in Figure~\ref{fig:intro}, \mbox{OmniMoGen} enables seamless and flexible motion generation across diverse objectives by merely adjusting the instruction. 
By empowering models to follow arbitrary instructions, this framework also opens up possibilities for creating new and imaginative motion generation tasks.

Unlike other generation methods, \mbox{OmniMoGen} features a unified and concise architecture, comprising an RVQ-VAE and a transformer model, as shown in Figure~\ref{fig:model}. 
Motions are encoded into discrete tokens like a foreign language by the RVQ-VAE, and then concatenated with text tokens as input to a unified autoregressive transformer for next-token prediction.
To train this architecture for free-form generation, we construct  \mbox{\textbf{X2Mo}}, the first large-scale multi-task motion generation dataset.
X2Mo consists of large-scale and high-quality interleaved text-motion instructions, benefiting from the automatic composition of motion captures.
It unifies in-context generation, motion editing, multi-turn editing, and reflection within free-form instruction following.

With the unified architecture and the interleaved instruction dataset, we train OmniMoGen in two stages, comprising multi-task supervised fine-tuning~(SFT) and GRPO-based reinforcement learning~(RL) for interleaved instruction following.
To facilitate a comprehensive evaluation of interleaved motion generation, we introduce a challenging benchmark named \mbox{\textbf{AnyContext}}. 
Each interleaved instruction in AnyContext is represented as a triplet consisting of a source motion, a reference motion, and a text description specifying motion attributes to be referenced.
We evaluate OmniMoGen and existing leading methods on \mbox{AnyContext}, where OmniMoGen achieves superior performance and exhibits emerging capabilities such as compositional editing, self-reflective generation, and knowledge-informed generation.
Furthermore, on existing text-to-motion and motion editing benchmarks, OmniMoGen achieves state-of-the-art results, outperforming the second-best method by 1.3\% on HumanML3D R@1 and by 1.4\% on Edited-to-Target R@1 on MotionFix~\cite{athanasiou2024motionfixtextdriven3dhuman}.
Extensive experimental results demonstrate that OmniMoGen achieves the next intelligent motion generation, capable of naturally generating human motions under a unified architecture following free-form instructions.

Our contributions are summarized as follows:

\begin{itemize}
    \item We introduce \mbox{OmniMoGen}, the first unified motion generation framework that can accommodate various motion generation tasks simply by adjusting the instruction.

    \item We construct a comprehensive motion generation dataset named \mbox{X2Mo}, consisting of interleaved text-motion instructions. And we further introduce \mbox{AnyContext} to evaluate the interleaved motion generation capability of mainstream methods.

    \item Our method achieves superior performance across \mbox{AnyContext}, text-to-motion, and motion editing benchmarks, exhibiting emerging capabilities such as compositional editing and self-reflective generation.
    
\end{itemize}

\section{Related Work}
\label{sec:related}

Human motion generation encompasses tasks such as text-to-motion generation, style editing, trajectory editing, motion inpainting, and in-betweening.
Existing methods typically treat these tasks as independent problems with distinct architectures and objectives.
For example, masked-transformer approaches~\cite{guo2023momaskgenerativemaskedmodeling, pinyoanuntapong2024mmmgenerativemaskedmotion,li2025lamplanguagemotionpretrainingmotion} improve generation quality and efficiency but remain constrained to text-to-motion and temporal inpainting due to predefined masked positions.
Diffusion-based methods~\cite{tevet2022humanmotiondiffusionmodel,athanasiou2024motionfixtextdriven3dhuman,jiang2025dynamicmotionblendingversatile} achieve high fidelity but rely on task-specific input formats, limiting them to isolated tasks.

Several works~\cite{guo2025motionlabunifiedhumanmotion,jiang2023motiongpt,wu2025mgmotionllmunifiedframeworkmotion,wu2024motionagentconversationalframeworkhuman} attempt to unify multiple tasks.
MotionLab~\cite{guo2025motionlabunifiedhumanmotion} adopts a Motion-Condition-Motion paradigm to support generation and editing, while MotionGPT~\cite{jiang2023motiongpt} focuses on tasks with simple contexts to unify understanding and generation.
However, these methods achieve partial unification and struggle to generalize to arbitrary motion tasks.
In contrast, OmniMoGen achieves strong performance across diverse generation tasks and further demonstrates emerging capabilities.

\section{Method}
\label{sec3:method}

\begin{figure*}[t]
\includegraphics[width=\linewidth]{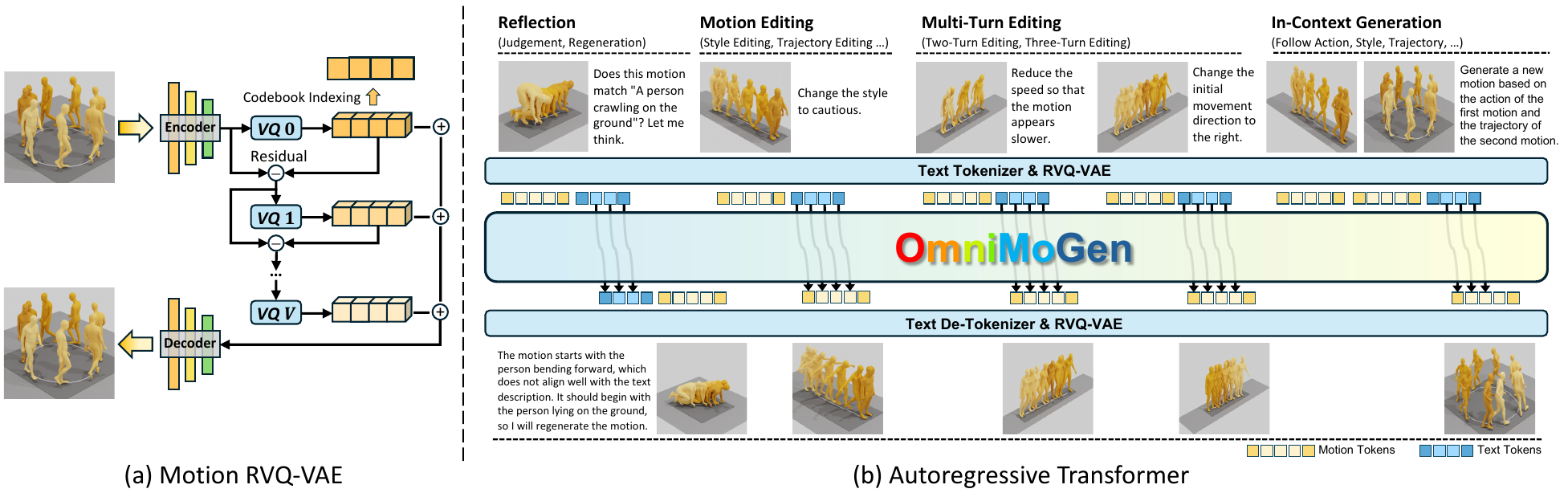}
\centering\caption{An overview of OmniMoGen, comprising (a) an RVQ-VAE and (b) an autoregressive transformer. Motions are encoded into discrete tokens like a foreign language by the RVQ-VAE, and then concatenated with text tokens as input to a unified autoregressive transformer for next-token prediction.}
\label{fig:model}
\end{figure*}

In this section, we first describe the large-scale interleaved text-motion instructions dataset, X2Mo~(§~\ref{sec3:dataset}). Then, we present the unified model architecture and two-stage training strategy~(§~\ref{sec3:model}). Finally, we introduce the AnyContext benchmark to evaluate interleaved motion generation~(§~\ref{sec3:bench}).

\subsection{X2Mo Dataset}
\label{sec3:dataset}
To unify various motion generation tasks, it is essential to train models on large-scale and diverse datasets. 
However, in the field of motion generation, such datasets are still lacking. 
In this work, we construct a large-scale unified motion generation dataset for the first time, referred to as the X2Mo dataset, consisting of 137K interleaved text-motion instructions with a unified format. 
We provide the statistics of X2Mo in Appendix~\textcolor{red}{A.1}. 

\subsubsection{Interleaved Text-Motion Instructions}

To enable unified learning across diverse motion generation tasks, we construct four types of interleaved text-motion instructions, namely in-context generation, motion editing, multi-turn editing, and reflection. 
These tasks enhance the understanding of interleaved text–motion instructions and support free-form motion generation under complex contexts.

\noindent 
\textbf{In-Context Generation.} 
In-context generation is well studied in image and text generation~\cite{wu2025omnigen2explorationadvancedmultimodal,tan2025ominicontrolminimaluniversalcontrol}, where models extract concepts from given examples and reproduce them in new outputs. 
However, such exploration remains rare in the field of motion generation.
To this end, we define an in-context generation task for the motion domain by decomposing motion into three fundamental elements: action, style, and trajectory. 
Action denotes the semantic category (e.g., jumping), style captures personalized variation (e.g., energetic), and trajectory specifies the spatial path (e.g., in place or a clockwise circle). 
As shown in Figure~\ref{fig:model}(b), these elements from multiple source motions are composited to generate novel motions that preserve contextual consistency.

\noindent
\textbf{Motion Editing.} 
The motion editing task focuses on modifying an existing motion based on natural language instructions while preserving its temporal dynamics and spatial coherence.
As shown in Figure~\ref{fig:model}(b), each sample contains a source motion, a target motion, and an editing instruction describing the modification, such as “\textit{Change the style to cautious}.”

\noindent
\textbf{Multi-Turn Editing.} 
The multi-turn editing task simulates real-world interactive motion refinement, where users iteratively modify an existing motion through natural language instructions. 
Each turn of editing introduces incremental changes, such as “\textit{Reduce the speed},” or “\textit{Change the initial movement direction to the right},” while preserving global temporal and spatial coherence.
As shown in Figure~\ref{fig:model}(b), each data sample contains an interleaved sequence of motions and editing instructions, enabling models to learn incremental refinement behaviors from natural language feedback.

\noindent
\textbf{Reflection.} Multimodal Large Language Models (MLLMs) have shown strong capabilities in test-time scaling and self-reflection~\cite{guo2025deepseek}, yet their application to motion generation remains limited. 
We explore integrating reflection mechanisms into motion generation and show that test-time scaling can further improve quality. 
As shown in Figure~\ref{fig:model}(b), the reflection data contain a chain of thought in which the model performs self-correction and progressive refinement, encouraging introspective reasoning that supports higher-quality motion generation.

\subsubsection{Data Construction}
\noindent
\textbf{Overview.} 
We introduce an automated pipeline for constructing interleaved text-motion instructions.
We first preprocess motion sequences into segments of up to 10 seconds at 20 fps.
Next, we compute similarities between segments using a motion retrieval model and connect similar segments to form a graph.
Finally, we collect interleaved text-motion pairs from the graph and use MLLMs to annotate them for specific tasks.

\noindent
\textbf{Motion Preprocessing.}
The AMASS dataset~\cite{mahmood2019amassarchivemotioncapture} provides a large collection of 3D human motion captures, and we adopt its motion sequences to ensure the high quality of X2Mo. 
Sequences longer than 10 seconds are segmented into action units following BABEL frame labels~\cite{punnakkal2021babelbodiesactionbehavior}, and all motions are downsampled to 20 fps to align with HumanML3D~\cite{guo2022generating}.
For a fair comparison, we discard motions included in the test sets of HumanML3D and MotionFix~\cite{athanasiou2024motionfixtextdriven3dhuman}.
We provide more details in Appendix~\textcolor{red}{A.2.1}.

\noindent
\textbf{Motion Graph Construction.}
We adopt the motion encoder from TMR~\cite{petrovich2023tmrtexttomotionretrievalusing} to map motion segments into a comparable embedding space and compute pairwise similarities.
An edge is added between two segments when their similarity exceeds 0.9, and to avoid cycles, we require the source node ID to be smaller than the target node ID.
Each segment is rendered into a video following~\cite{petrovich2022temosgeneratingdiversehuman}, and \mbox{Gemini-2.5-Pro} is used to annotate fine-grained action information, such as \textit{action type}, \textit{body part}, \textit{style}, \textit{duration}, and \textit{trajectory}.
We provide a set of predefined options for each type of action information to avoid overly diverse outputs.
If two segments share identical action lists, their edge is discarded. 
The remaining edges form a directed acyclic motion graph.
We provide more details in Appendix~\textcolor{red}{A.2.2}.

\noindent
\textbf{Interleaving from Motion Graph.}
Next, we construct four types of interleaved text-motion instructions from the motion graph.
Specifically, \textbf{1)~in-context generation}: we identify converging substructures where multiple source nodes share the same target node. 
If the target action list can be covered by those of the source nodes, we construct an in-context instruction, such as “\textit{concatenate walking in motion A with turning in motion B}.” 
\textbf{2)~motion editing}: for each edge, we design an editing instruction based on the differences between the action lists of the source and target segments, e.g., if the target includes an extra kick, the instruction becomes “\textit{add a kicking at the end}.”
\textbf{3)~multi-turn editing}: to simulate multi-turn scenarios, we sample continuous paths on the motion graph. 
We treat the first node as the initial motion, and iteratively generate editing instructions that transform the current motion into the next node, with each instruction derived from the action-list differences between consecutive nodes.
\textbf{4)~reflection}: for each edge, we take the caption of the target node as the input description and construct positive and negative pairs. 
Positive samples use the target motion (caption and motion aligned), while negative samples use the source motion (caption and motion misaligned). 
Based on these aligned and misaligned pairs, we employ MLLMs to generate a detailed reasoning process explaining why the motion matches or mismatches the caption.
We provide more details in Appendix~\textcolor{red}{A.2.3}.

\subsection{OmniMoGen}
\label{sec3:model}
We introduce a unified architecture, comprising an RVQ-VAE and an autoregressive transformer. With the unified architecture and the interleaved instruction dataset, we train OmniMoGen in two stages, comprising multi-task SFT and GRPO-based RL for interleaved instruction following. 

\subsubsection{Model Architecture.} 
Unlike existing task-specific designs that rely on heterogeneous modules, \mbox{OmniMoGen} adopts a unified architecture with two core components: an RVQ-VAE motion tokenizer and an autoregressive transformer, as illustrated in Figure~\ref{fig:model}. 

Since VQ-VAE suffers from limited reconstruction fidelity, we adopt an RVQ-VAE that progressively refines quantization through multiple residual codebooks. 
Given an input motion $\mathbf{M} = [\mathbf{m}_1, \dots, \mathbf{m}_T]$, the encoder $\mathcal{E}$ maps it to latent features $\mathbf{Z}$, which are quantized by $L$ hierarchical codebooks $\{\mathcal{C}^{(1)}, \dots, \mathcal{C}^{(L)}\}$. 
Each level refines the residual from the previous one:
\begin{equation}
\hat{\mathbf{z}}_t = \sum_{l=1}^{L} q^{(l)}\big(\mathbf{r}_t^{(l-1)}\big), \quad 
\mathbf{r}_t^{(l)} = \mathbf{r}_t^{(l-1)} - q^{(l)}\big(\mathbf{r}_t^{(l-1)}\big).
\end{equation}
The decoder $\mathcal{D}$ reconstructs the motion $\hat{\mathbf{M}} = \mathcal{D}(\hat{\mathbf{Z}})$, and the model is trained with standard reconstruction and commitment losses. 
After training, motion is represented by a stack of discrete indices $(v_t^{(1)}, \dots, v_t^{(L)})$, forming the tokenized motion $\mathbf{v}$.

Following tokenization, we denote the motion token vocabulary as $\mathcal{V}_M = \{\texttt{<Motion\_i>}\}_{i=1}^{K}$ and introduce \texttt{<Motion>} and \texttt{</Motion>} tokens to mark sequence boundaries. 
These tokens are appended to the transformer’s textual vocabulary, forming a unified token space for joint modeling. 
During generation, the transformer receives an interleaved sequence of text and motion tokens $\mathbf{s} = [x_1, \dots, x_N, v_1, \dots, v_T]$ and predicts the next token by maximizing:
\begin{equation}
\mathcal{L}_{\text{LLM}} = \sum_{t} \log p_{\psi}(s_t \mid s_{<t}),
\end{equation}
where $p_{\psi}$ is parameterized by the transformer.

\subsubsection{Supervised Fine-Tuning.}
We first conduct supervised fine-tuning~(SFT) on the X2Mo dataset, enabling OmniMoGen to follow interleaved instructions. 
Each motion is encoded into multi-layer discrete tokens by the RVQ-VAE, where tokens from all quantization layers are flattened into a single sequence before being concatenated with the corresponding text tokens.
This flattened interleaved sequence serves as the input for the transformer model, which autoregressively predicts the next token conditioned on the preceding multimodal context.

Formally, given a training sequence $\mathbf{x} = [x_1, x_2, \dots, x_T]$ interleaved with text and flattened motion tokens, the model is optimized by minimizing the negative log-likelihood:
\begin{equation}
\mathcal{L}_{\text{SFT}} = -\sum_{t=1}^{T} \log p_\theta(x_t \mid x_{<t}),
\end{equation}
where $\theta$ denotes the model parameters. 
We apply teacher forcing and compute cross-entropy loss over both text and motion tokens, guiding the model to capture fine-grained token dependencies across modalities and quantization levels.

Through the SFT stage, OmniMoGen learns to follow interleaved text-motion instructions and generate coherent motions aligned with instruction intent. 

\subsubsection{Reinforcement Learning.} 
The SFT stage provides the foundation for interleaved instruction following, while the RL stage further enhances generalization and reasoning capabilities in free-form motion generation.

To optimize OmniMoGen beyond imitation learning, we perform reinforcement learning~(RL) on the X2Mo dataset. Given a generated motion $\hat{M}$, the total reward is defined as:
\begin{equation}
\mathcal{R} = 
\lambda_{\text{sem}} \, \mathcal{R}_{\text{sem}} 
+ 
\lambda_{\text{phy}} \, \mathcal{R}_{\text{phy}},
\end{equation}
where $\lambda_{\text{sem}}$ and $\lambda_{\text{phy}}$ balance the two rewards.

\noindent
\textbf{Semantic Correctness Reward.}
Each instruction in X2Mo corresponds to a ground-truth target motion, allowing a retrieval-based reward that measures how well the generated motion aligns with the target. For $N$ generated motions $\{\hat{M}_i\}$ and their instructions $\{T_i\}$, we use a pretrained motion-text encoder $f(\cdot)$ to compute feature similarity:
\begin{align}
\mathcal{R}_{\text{sem}}^{(i)} 
&= 
\frac{
\exp\!\left(
\mathrm{sim}\!\left(f_m(\hat{M}_i), f_t(T_i)\right)/\tau
\right)
}{
\sum_{j=1}^{N}
\exp\!\left(
\mathrm{sim}\!\left(f_m(\hat{M}_i), f_t(T_j)\right)/\tau
\right)
},
\label{eq:semantic_reward}
\end{align}
where $\mathrm{sim}(\cdot,\cdot)$ denotes cosine similarity and $\tau$ is a temperature parameter.

\noindent
\textbf{Physical Plausibility Reward.}
To ensure physical realism, we introduce a skating-based reward that penalizes foot sliding. Let $p_i^t(x,y)$ denote the horizontal position of foot joint 
$i \in \{\text{L/R Ankle},\text{L/R Toe}\}$ and $c_i^t$ its contact indicator. The reward is:
\begin{align}
\mathcal{R}_{\text{phy}} 
&= 
- \frac{1}{T} 
\sum_{t}
\sum_{i \in \{\text{LA,LT,RA,RT}\}} 
c_i^t 
\left\| 
\dot{p}_i^t(x,y) 
\right\|_2^2.
\label{eq:physical_reward}
\end{align}

We employ the GRPO algorithm to maximize the expected cumulative reward. For sampled sequences $\{x_i\}$ with rewards $\{\mathcal{R}_i\}$ and baseline $b$, the policy gradient is:
\begin{equation}
\nabla_{\theta} \mathcal{J}(\theta)
=
\mathbb{E}_{x \sim p_{\theta}}
\!\left[
\nabla_{\theta} \log p_{\theta}(x)
\cdot
\left(
\mathcal{R}(x) - b
\right)
\right].
\label{eq:policy_gradient}
\end{equation}
Following GRPO, we adopt a clipped surrogate objective:
\begin{align}
\mathcal{L}_{\text{GRPO}} 
&=
\mathbb{E}_t 
\Big[
\min\!\Big(
r_t(\theta) A_t,\;
\mathrm{clip}\!\left(r_t(\theta), 1-\epsilon, 1+\epsilon\right) A_t
\Big)
\nonumber \\
&\quad
- \beta 
D_{\mathrm{KL}}\!\left[ 
p_{\theta_{\text{old}}} \; \| \; p_{\theta}
\right]
\Big],
\label{eq:grpo_loss}
\end{align}
where 
$r_t(\theta) = 
\dfrac{
p_{\theta}(x_t|x_{<t})
}{
p_{\theta_{\text{old}}}(x_t|x_{<t})
}$ and $\beta$ controls KL regularization.

Through RL, OmniMoGen goes beyond imitation learning and improves reasoning over interleaved text-motion contexts.

\subsection{AnyContext Benchmark}
\label{sec3:bench}
In traditional motion generation benchmarks, each test case contains a simple and task-specific context, such as a single text description. 
Under this setting, methods may achieve strong performance on specific tasks even if they fail to handle arbitrary contexts. 
However, such performance does not transfer well to real-world scenarios with interleaved text-motion contexts, making it difficult to reflect their true generation capability. 
To this end, we introduce a new benchmark, \mbox{AnyContext}, which evaluates interleaved motion generation.

\begin{figure}
    \centering
    \includegraphics[width=\linewidth]{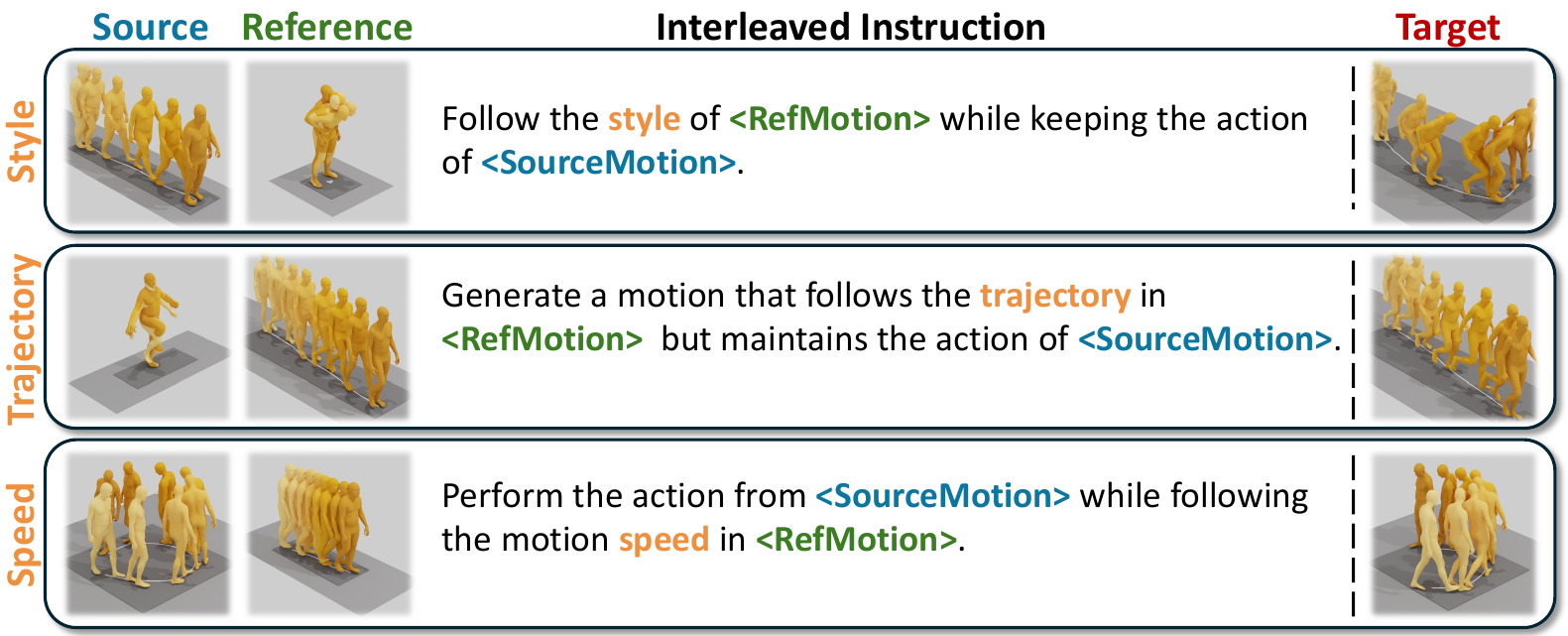}
    \caption{Task types and interleaved text–motion instruction formats in AnyContext}
    \label{fig:bench}
\end{figure}

\subsubsection{Task Categorization}
Each test case consists of a source motion, a reference motion, and a text description. 
The source motion defines the action type, while the text description specifies which attribute of the reference motion should be referenced. 
Based on these attributes, we categorize the tasks in \mbox{AnyContext} into three types: \textit{1)~Style-based}, \textit{2)~Trajectory-based}, and \textit{3)~Speed-based} generation.
In Figure~\ref{fig:bench}, we present examples for each category. We provide more details in Appendix~\textcolor{red}{B.1}.

\subsubsection{Evaluation Metrics}
We adopt a retrieval accuracy metric and a physical plausibility metric to evaluate interleaved motion generation.

\noindent
\textbf{Retrieval Accuracy.}
We use a retrieval-based metric similar to~\cite{guo2022generating,athanasiou2024motionfixtextdriven3dhuman} to assess semantic alignment between generated and target motions. 
All motions are encoded using TMR~\cite{petrovich2023tmrtexttomotionretrievalusing} features, and pairwise similarities are computed against a randomly sampled gallery. 
For each task category in \mbox{AnyContext}, we report R@1, R@3, and AvgR under 32-sample random gallery evaluation.

\noindent
\textbf{Physical Plausibility.}
We also report a physical plausibility score that reflects the real-world quality of generated motions~\cite{zhao2023synthesizing, jiang2025dynamic}. 
For each frame, we compute a foot contact score over L/R Ankle and L/R Toe:
\begin{equation}
s_{\text{contact}}
= 
\exp\!\left(-\left(|z_{\min}| - \tau_h\right)^{+}\right)
\cdot
\exp\!\left(-\left(\|v_{\min}\|_2 - \tau_v\right)^{+}\right),
\end{equation}
where $z_{\min}$ and $v_{\min}$ denote the minimum foot--ground distance and horizontal velocity, $(\cdot)^{+}$ is the ReLU operator, and $\tau_h = 0.05\text{m}$, $\tau_v = 0.075\text{m/s}$. 
A higher score indicates more stable grounding and less sliding. We report the sequence-averaged score as the physical plausibility metric, reflecting artifacts such as foot floating and skating.

\section{Experiments}
\label{sec4:experiments}
In this section, we conduct experiments to validate the effectiveness of OmniMoGen. 
We evaluate OmniMoGen on three representative tasks, including text-to-motion generation (§~\ref{sec4:t2m}), motion editing (§~\ref{sec4:edit}), and interleaved motion generation on AnyContext (§~\ref{sec4:bench}).
Finally, we demonstrate the emerging capabilities and provide in-depth analyses.

\subsection{Experimental Setup}
\label{sec4:setup}

\noindent 
\textbf{Settings.} 
For the motion RVQ-VAE, we follow the network structure and training strategy in MoMask~\cite{guo2023momaskgenerativemaskedmodeling}. It consists of 6 quantization layers, where each layer’s codebook contains 512 code vectors, each of 512 dimensions. The quantization dropout ratio $q$ is set to 0.2. 
For the autoregressive transformer, we use Gemma2-2B~\cite{gemmateam2024gemma2improvingopen}, a lightweight open-source LLM from Google.
Benefiting from training on the X2Mo-Reflection subset, we introduce a variant named \textbf{\mbox{OmniMoGen-Think}}. In the think mode, OmniMoGen is deliberately guided to perform self-reflection after generating a motion, and subsequently refine the motion based on the reflection.
Considering the time cost, we set the maximum number of reflection rounds to 3.
All experiments are conducted with NVIDIA A100 80G GPUs.

\noindent 
\textbf{Baselines and Benchmarks.} 
We evaluate text-to-motion generation on HumanML3D~\cite{guo2022generating}, motion editing on MotionFix~\cite{athanasiou2024motionfixtextdriven3dhuman}, and interleaved motion generation on AnyContext. 
We compare our method against 21 motion generation baselines grouped into three categories: 
1)~Autoregressive-based methods, such as \mbox{T2M-GPT}~\cite{zhang2023t2mgptgeneratinghumanmotion}, and \mbox{MotionGPT}; 
2)~Diffusion-based methods, such as \mbox{MLD}~\cite{chen2023executingcommandsmotiondiffusion}, and \mbox{MotionReFit}~\cite{jiang2025dynamicmotionblendingversatile}; 
and 3)~Hybrid architectures, such as \mbox{MotionGPT3}~\cite{zhu2025motiongpt3humanmotionsecond}.

\subsection{Text-to-Motion Generation}
\label{sec4:t2m}

\noindent
\textbf{Metric Evaluation.}
We evaluate the text-to-motion generation capability of \mbox{OmniMoGen} on the \mbox{HumanML3D}~\cite{guo2022generating} test set, comparing the performance of OmniMoGen with mainstream motion generation methods, as shown in Table~\ref{tab:t2m}.
The evaluations are conducted 20 times to obtain a 95\% confidence interval.
Notably, OmniMoGen achieves competitive performance across metrics such as RPrecision, FID, and MMDist, which measure motion accuracy.
Among multi-task baselines, OmniMoGen achieves \textbf{state-of-the-art results} on these metrics by extending the inference time, outperforming the second-best method by 1.3\% on R@1, reducing FID by 0.13 and MMDist by 2.29\%.

\begin{table*}[h!]
\centering
\renewcommand{\arraystretch}{0.7}

\caption{Comparison with existing motion generation methods. The evaluations are conducted 20 times to obtain a 95\% confidence interval~(±). Best results are highlighted in \textbf{bold} and the second best in \underline{underline}. }
\resizebox{\textwidth}{!}{
\begin{tabular}{llccccccc}
\toprule
\multirow{2}{*}{\centering \textbf{Types}} &
\multirow{2}{*}{\textbf{Methods}} & \multicolumn{3}{c}{\textbf{RPrecision}$\uparrow$} & \multirow{2}{*}{\textbf{FID}$\downarrow$} & \multirow{2}{*}{\textbf{MMDist}$\downarrow$} & \multirow{2}{*}{\textbf{Diversity}$\uparrow$} & \multirow{2}{*}{\textbf{MModality}$\uparrow$} \\
\cmidrule(lr){3-5}
 &  & Top-1 & Top-2 & Top-3 &  &  &  &  \\
\midrule
& Real & $0.511^{\text{\scriptsize $\pm$.003}}$ & $0.703^{\text{\scriptsize $\pm$.003}}$ & $0.797^{\text{\scriptsize $\pm$.002}}$ & $0.002^{\text{\scriptsize $\pm$.000}}$ & $2.974^{\text{\scriptsize $\pm$.008}}$ & $9.503^{\text{\scriptsize $\pm$.065}}$ & - \\
\midrule
\multirow{9}{*}{\centering \textit{Text-to-Motion Only}}  & TM2T~\cite{guo2022tm2t} & $0.424^{\text{\scriptsize $\pm$.003}}$ & $0.618^{\text{\scriptsize $\pm$.003}}$ & $0.729^{\text{\scriptsize $\pm$.002}}$ & $1.501^{\text{\scriptsize $\pm$.017}}$ & $3.467^{\text{\scriptsize $\pm$.011}}$ & $8.589^{\text{\scriptsize $\pm$.076}}$ & $\underline{2.424}^{\text{\scriptsize $\pm$.093}}$ \\
& T2M~\cite{guo2022generating} & $0.457^{\text{\scriptsize $\pm$.002}}$ & $0.639^{\text{\scriptsize $\pm$.003}}$ & $0.740^{\text{\scriptsize $\pm$.003}}$ & $1.067^{\text{\scriptsize $\pm$.002}}$ & $3.340^{\text{\scriptsize $\pm$.008}}$ & $9.188^{\text{\scriptsize $\pm$.082}}$ & $2.090^{\text{\scriptsize $\pm$.083}}$ \\
& MotionDiffuse~\cite{zhang2022motiondiffusetextdrivenhumanmotion} & $0.491^{\text{\scriptsize $\pm$.001}}$ & $0.681^{\text{\scriptsize $\pm$.001}}$ & $0.782^{\text{\scriptsize $\pm$.001}}$ & $0.630^{\text{\scriptsize $\pm$.001}}$ & $3.113^{\text{\scriptsize $\pm$.001}}$ & $9.410^{\text{\scriptsize $\pm$.049}}$ & $1.553^{\text{\scriptsize $\pm$.042}}$ \\
& MDM~\cite{tevet2022humanmotiondiffusionmodel} & $0.320^{\text{\scriptsize $\pm$.005}}$ & $0.498^{\text{\scriptsize $\pm$.004}}$ & $0.611^{\text{\scriptsize $\pm$.007}}$ & $0.544^{\text{\scriptsize $\pm$.044}}$ & $5.566^{\text{\scriptsize $\pm$.027}}$ & $9.559^{\text{\scriptsize $\pm$.086}}$ & $\mathbf{2.799}^{\text{\scriptsize $\pm$.072}}$ \\
& MLD~\cite{chen2023executingcommandsmotiondiffusion} & $0.481^{\text{\scriptsize $\pm$.003}}$ & $0.673^{\text{\scriptsize $\pm$.003}}$ & $0.772^{\text{\scriptsize $\pm$.002}}$ & $0.473^{\text{\scriptsize $\pm$.013}}$ & $3.196^{\text{\scriptsize $\pm$.010}}$ & $9.724^{\text{\scriptsize $\pm$.082}}$ & $2.413^{\text{\scriptsize $\pm$.079}}$ \\
& T2M-GPT~\cite{zhang2023t2mgptgeneratinghumanmotion} & $0.491^{\text{\scriptsize $\pm$.003}}$ & $0.680^{\text{\scriptsize $\pm$.003}}$ & $0.775^{\text{\scriptsize $\pm$.002}}$ & $0.116^{\text{\scriptsize $\pm$.002}}$ & $3.118^{\text{\scriptsize $\pm$.011}}$ & $\underline{9.761}^{\text{\scriptsize $\pm$.081}}$ & $1.856^{\text{\scriptsize $\pm$.011}}$ \\
& Motion-R1~\cite{ouyang2025motionr1chainofthoughtreasoningreinforcement} & $0.515^{\text{\scriptsize $\pm$.003}}$ & $\underline{0.719}^{\text{\scriptsize $\pm$.002}}$ & $\underline{0.818}^{\text{\scriptsize $\pm$.002}}$ & $0.201^{\text{\scriptsize $\pm$.004}}$ & $\underline{2.854}^{\text{\scriptsize $\pm$.010}}$ & $\mathbf{10.026}^{\text{\scriptsize $\pm$.075}}$ & $2.317^{\text{\scriptsize $\pm$.105}}$ \\
& MoMask~\cite{guo2023momaskgenerativemaskedmodeling} & $\underline{0.521}^{\text{\scriptsize $\pm$.002}}$ & $0.713^{\text{\scriptsize $\pm$.002}}$ & $0.807^{\text{\scriptsize $\pm$.002}}$ & $\mathbf{0.045}^{\text{\scriptsize $\pm$.002}}$ & $2.958^{\text{\scriptsize $\pm$.008}}$ & $9.620^{\text{\scriptsize $\pm$.064}}$ & $1.241^{\text{\scriptsize $\pm$.040}}$ \\
& SALAD~\cite{hong2025saladskeletonawarelatentdiffusion} & $\mathbf{0.581}^{\text{\scriptsize $\pm$.003}}$ & $\mathbf{0.769}^{\text{\scriptsize $\pm$.003}}$ & $\mathbf{0.857}^{\text{\scriptsize $\pm$.002}}$ & $\underline{0.076}^{\text{\scriptsize $\pm$.002}}$ & $\mathbf{2.649}^{\text{\scriptsize $\pm$.009}}$ & $9.696^{\text{\scriptsize $\pm$.096}}$ & $1.751^{\text{\scriptsize $\pm$.062}}$ \\
\midrule
\multirow{9}{*}{\centering \textit{Multi-Task}} & MotionLLM~\cite{chen2024motionllmunderstandinghumanbehaviors} & $0.515^{\text{\scriptsize $\pm$.004}}$ & - & $0.801^{\text{\scriptsize $\pm$.004}}$ & $0.230^{\text{\scriptsize $\pm$.009}}$ & $2.967^{\text{\scriptsize $\pm$.020}}$ & $9.908^{\text{\scriptsize $\pm$.102}}$ & - \\
& MotionGPT~\cite{zhang2024motiongpt} & $0.364^{\text{\scriptsize $\pm$.005}}$ & $0.533^{\text{\scriptsize $\pm$.003}}$ & $0.629^{\text{\scriptsize $\pm$.004}}$ & $0.805^{\text{\scriptsize $\pm$.002}}$ & $3.914^{\text{\scriptsize $\pm$.013}}$ & $\mathbf{9.972}^{\text{\scriptsize $\pm$.026}}$ & $\mathbf{2.473}^{\text{\scriptsize $\pm$.041}}$ \\
& MG-MotionLLM~\cite{wu2025mgmotionllmunifiedframeworkmotion} & $0.516^{\text{\scriptsize $\pm$.002}}$ & $0.706^{\text{\scriptsize $\pm$.002}}$ & $0.802^{\text{\scriptsize $\pm$.003}}$ & $0.303^{\text{\scriptsize $\pm$.010}}$ & $2.952^{\text{\scriptsize $\pm$.009}}$ & $\underline{9.960}^{\text{\scriptsize $\pm$.073}}$ & $2.125^{\text{\scriptsize $\pm$.159}}$ \\
& MotionGPT~\cite{jiang2023motiongpt} & $0.492^{\text{\scriptsize $\pm$.003}}$ & $0.681^{\text{\scriptsize $\pm$.003}}$ & $0.778^{\text{\scriptsize $\pm$.002}}$ & $0.232^{\text{\scriptsize $\pm$.008}}$ & $3.096^{\text{\scriptsize $\pm$.008}}$ & $9.528^{\text{\scriptsize $\pm$.071}}$ & $2.008^{\text{\scriptsize $\pm$.084}}$ \\
& MotionChain~\cite{jiang2024motionchainconversationalmotioncontrollers} & $0.504^{\text{\scriptsize $\pm$.003}}$ & $0.695^{\text{\scriptsize $\pm$.003}}$ & $0.790^{\text{\scriptsize $\pm$.003}}$ & $0.248^{\text{\scriptsize $\pm$.009}}$ & $3.033^{\text{\scriptsize $\pm$.010}}$ & $9.470^{\text{\scriptsize $\pm$.075}}$ & $1.715^{\text{\scriptsize $\pm$.066}}$ \\
& MotionGPT-2~\cite{wang2024motiongpt2generalpurposemotionlanguagemodel} & $0.496^{\text{\scriptsize $\pm$.002}}$ & $0.691^{\text{\scriptsize $\pm$.003}}$ & $0.782^{\text{\scriptsize $\pm$.004}}$ & $0.191^{\text{\scriptsize $\pm$.004}}$ & $3.080^{\text{\scriptsize $\pm$.013}}$ & $9.860^{\text{\scriptsize $\pm$.026}}$ & $\underline{2.137}^{\text{\scriptsize $\pm$.022}}$ \\
& MotionGPT3~\cite{zhu2025motiongpt3humanmotionsecond} & $\underline{0.543}^{\text{\scriptsize $\pm$.003}}$ & $\underline{0.735}^{\text{\scriptsize $\pm$.002}}$ & $\underline{0.828}^{\text{\scriptsize $\pm$.002}}$ & $0.217^{\text{\scriptsize $\pm$.010}}$ & $2.793^{\text{\scriptsize $\pm$.007}}$ & $9.662^{\text{\scriptsize $\pm$.072}}$ & $1.366^{\text{\scriptsize $\pm$.046}}$ \\
\cmidrule(lr){2-9}
& \textbf{OmniMoGen} & $0.525^{\text{\scriptsize $\pm$.003}}$ &
$0.719^{\text{\scriptsize $\pm$.003}}$ &
$0.814^{\text{\scriptsize $\pm$.002}}$ &
$\underline{0.093}^{\text{\scriptsize $\pm$.005}}$ &
$\underline{2.761}^{\text{\scriptsize $\pm$.012}}$ &
$9.692^{\text{\scriptsize $\pm$.061}}$ & 
$1.464^{\text{\scriptsize $\pm$.079}}$ \\
& \textbf{OmniMoGen-Think} & $\mathbf{0.550^{\text{\scriptsize $\pm$.004}}}$ &
$\mathbf{0.742^{\text{\scriptsize $\pm$.004}}}$ &
$\mathbf{0.834^{\text{\scriptsize $\pm$.003}}}$ &
$\mathbf{0.061^{\text{\scriptsize $\pm$.010}}}$ &
$\mathbf{2.729^{\text{\scriptsize $\pm$.010}}}$ &
$9.730^{\text{\scriptsize $\pm$.082}}$ & 
$1.673^{\text{\scriptsize $\pm$.134}}$ \\
\bottomrule
\end{tabular}
}

\label{tab:t2m}
\end{table*}

\begin{figure}
    \centering
    \includegraphics[width=\linewidth]{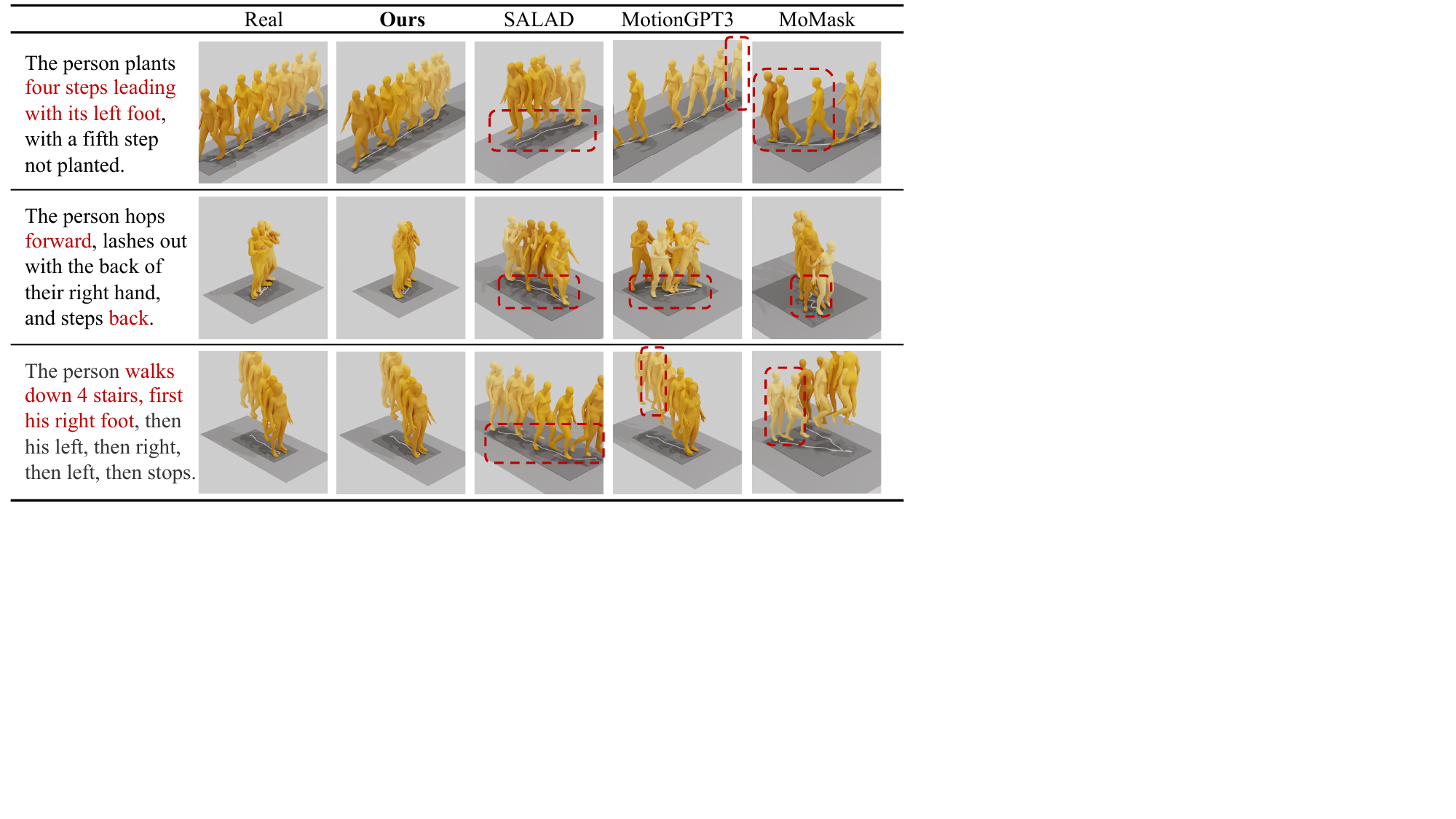}
    \caption{Qualitative comparison of text-to-motion generation on \mbox{HumanML3D}. The red words and boxes highlight the misaligned motions.}
    \vspace{-1em}
    \label{fig:q_t2m}
\end{figure}

\noindent
\textbf{Qualitative Results.} 
As shown in Figure~\ref{fig:q_t2m}, we conduct a qualitative comparison between OmniMoGen and leading text-to-motion generation methods, including SALAD~\cite{hong2025saladskeletonawarelatentdiffusion}, MotionGPT3~\cite{zhu2025motiongpt3humanmotionsecond}, and MoMask~\cite{guo2023momaskgenerativemaskedmodeling}.
Although SALAD~\cite{hong2025saladskeletonawarelatentdiffusion} achieves impressive quantitative metrics, it shows poor control over the distance between the feet and the ground.
MotionGPT3~\cite{zhu2025motiongpt3humanmotionsecond} often confuses ``\textit{left foot}'' and ``\textit{right foot}.''
MoMask~\cite{guo2023momaskgenerativemaskedmodeling} often generates motions with inaccurate trajectories.
In comparison, OmniMoGen generates high-quality motions faithful to the text description.

\subsection{Motion Editing}
\label{sec4:edit}

\noindent
\textbf{Metric Evaluation.}
We evaluate the motion editing capability of \mbox{OmniMoGen} on the \mbox{MotionFix}~\cite{athanasiou2024motionfixtextdriven3dhuman} test set. 
\mbox{OmniMoGen} employs the 263-dimensional motion representation defined in HumanML3D~\cite{guo2022generating}, whereas the MotionFix dataset uses the SMPL format. 
For evaluation on the MotionFix dataset, we first compute 3D joint coordinates from the 263-dimensional vectors and then estimate SMPL parameters using inverse kinematics.
Additionally, we unify the fps to 30, consistent with the setting used in MotionFix.
As shown in Table~\ref{tab:edit}, \mbox{OmniMoGen} achieves \textbf{the best performance across all metrics}, significantly outperforming prior leading methods~\cite{li2025simmotionedittextbasedhumanmotion,jiang2025dynamicmotionblendingversatile,athanasiou2024motionfixtextdriven3dhuman}.
Specifically, OmniMoGen outperforms the second-best method by 5.3\% on Edited-to-Source R@1 and 1.4\% on Edited-to-Target R@1.
Notably, although OmniMoGen introduces errors during the conversion to the SMPL format, it still outperforms other methods in overall performance.

\begin{table}[t!]
\centering

\caption{Quantitative comparison on MotionFix.}
\resizebox{\linewidth}{!}{
\begin{tabular}{lcccccccc}
\toprule
\multirow{2}{*}{\textbf{Methods}} & \multicolumn{4}{c}{\textbf{Edited-to-Source Retrieval}} & \multicolumn{4}{c}{\textbf{Edited-to-Target Retrieval}} \\
\cmidrule(lr){2-5} \cmidrule(lr){6-9}
 & R@1$\uparrow$ & R@2$\uparrow$ & R@3$\uparrow$ & AvgR$\downarrow$ & R@1$\uparrow$ & R@2$\uparrow$ & R@3$\uparrow$ & AvgR$\downarrow$ \\
\midrule
Real Data & 74.01 & 84.52 & 89.91 & 2.03 & 100.0 & 100.0 & 100.0 & 1.00 \\
MDM-BP~\cite{athanasiou2024motionfixtextdriven3dhuman} & 61.28 & 69.55 & 73.99 & 4.21 & 39.10 & 50.09 & 54.84 & 6.46 \\
TMED~\cite{athanasiou2024motionfixtextdriven3dhuman} & 71.77 & 84.07 & 89.52 & 1.96 & 62.90 & 76.51 & 83.06 & 2.71 \\
MotionReFit~\cite{jiang2025dynamicmotionblendingversatile} & 83.47 & 90.42 & 92.84 & 1.73 & 66.33 & 80.05 & 84.98 & 2.64 \\
SimMotionEdit~\cite{li2025simmotionedittextbasedhumanmotion} & 72.71 & 83.54 & 87.50 & / & \underline{70.62} & \underline{82.92} & \underline{88.12} & \underline{2.38} \\
\midrule
\textbf{OmniMoGen} & \underline{85.41} & \underline{92.13} & \underline{93.17} & \underline{1.60} & 68.33 & 81.85 & 87.42 & 2.47 \\
\textbf{OmniMoGen-Think} & \textbf{87.87} & \textbf{93.02} & \textbf{94.57} & \textbf{1.45} & \textbf{71.59} & \textbf{83.50} & \textbf{88.80} & \textbf{2.31} \\
\bottomrule
\end{tabular}
}

\label{tab:edit}
\end{table}

\begin{figure}
    \centering
    \includegraphics[width=\linewidth]{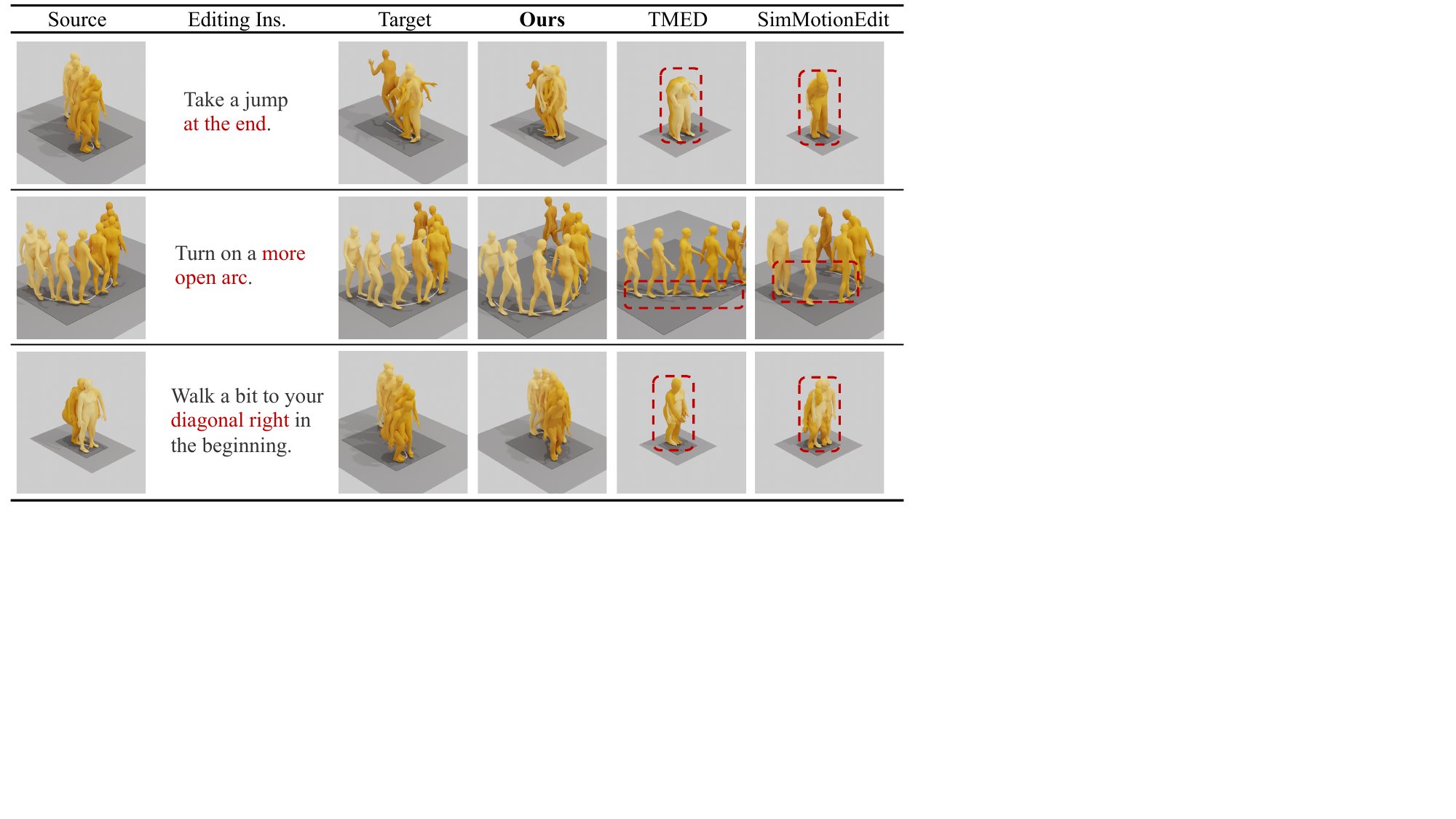}
    \caption{Qualitative comparison of motion editing on \mbox{MotionFix}. The red words and boxes highlight the misaligned motions.}
    \vspace{-1em}
    \label{fig:q_edit}
\end{figure}

\noindent
\textbf{Qualitative Results.} 
We conduct a qualitative comparison of OmniMoGen and leading motion editing methods, including TMED, and SimMotionEdit~\cite{li2025simmotionedittextbasedhumanmotion}, in Figure~\ref{fig:q_edit}.
We can observe that TMED and SimMotionEdit often struggle to precisely interpret edit instructions.
In comparison, OmniMoGen generates edited motions that closely match the target motion while making minimal unnecessary changes to the source motion.

\subsection{Comparison on AnyContext}
\label{sec4:bench}
\begin{table*}[t!]
\centering

\caption{Quantitative comparison on AnyContext across three task types: Style, Trajectory, and Speed. 
The evaluations are conducted 20 times to obtain a 95\% confidence interval~(±). 
Best results are highlighted in \textbf{bold}.
``Physical'' measures the physical plausibility metric.}
\resizebox{\textwidth}{!}{
\begin{tabular}{lcccccccccccc}
\toprule
\multirow{2}{*}{\textbf{Methods}} 
& \multicolumn{4}{c}{\textbf{Style-based}} 
& \multicolumn{4}{c}{\textbf{Trajectory-based}} 
& \multicolumn{4}{c}{\textbf{Speed-based}} \\
\cmidrule(lr){2-5} \cmidrule(lr){6-9} \cmidrule(lr){10-13}
 & R@1$\uparrow$ & R@3$\uparrow$ & AvgR$\downarrow$ & Physical$\uparrow$
 & R@1$\uparrow$ & R@3$\uparrow$ & AvgR$\downarrow$ & Physical$\uparrow$
 & R@1$\uparrow$ & R@3$\uparrow$ & AvgR$\downarrow$ & Physical$\uparrow$ \\
\midrule
Real Data 
& $100.0^{\scriptsize \pm.000}$ & $100.0^{\scriptsize \pm.000}$ & $1.0^{\scriptsize \pm.000}$ & $0.98^{\scriptsize \pm.000}$
& $100.0^{\scriptsize \pm.000}$ & $100.0^{\scriptsize \pm.000}$ & $1.0^{\scriptsize \pm.000}$ & $0.99^{\scriptsize \pm.000}$
& $100.0^{\scriptsize \pm.000}$ & $100.0^{\scriptsize \pm.000}$ & $1.0^{\scriptsize \pm.000}$ & $0.98^{\scriptsize \pm.000}$ \\
\midrule
AttT2M~\cite{zhong2023attt2mtextdrivenhumanmotion} 
& $13.6^{\scriptsize \pm.742}$ & $18.4^{\scriptsize \pm1.219}$ & $17.9^{\scriptsize \pm.130}$ & $0.88^{\scriptsize \pm.009}$
& $17.2^{\scriptsize \pm1.532}$ & $21.7^{\scriptsize \pm.904}$ & $13.4^{\scriptsize \pm.151}$ & $0.88^{\scriptsize \pm.010}$
& $21.3^{\scriptsize \pm.653}$ & $26.2^{\scriptsize \pm1.447}$ & $10.1^{\scriptsize \pm.189}$ & $0.89^{\scriptsize \pm.012}$ \\
T2M-GPT~\cite{zhang2023t2mgptgeneratinghumanmotion}
& $16.4^{\scriptsize \pm1.102}$ & $20.4^{\scriptsize \pm.947}$ & $14.1^{\scriptsize \pm.058}$ & $0.89^{\scriptsize \pm.008}$
& $14.7^{\scriptsize \pm.803}$ & $19.6^{\scriptsize \pm1.794}$ & $18.3^{\scriptsize \pm.196}$ & $0.88^{\scriptsize \pm.005}$
& $25.1^{\scriptsize \pm.641}$ & $29.4^{\scriptsize \pm1.224}$ & $9.6^{\scriptsize \pm.174}$ & $0.87^{\scriptsize \pm.009}$ \\
MotionGPT~\cite{zhang2024motiongpt}
& $19.9^{\scriptsize \pm1.537}$ & $24.4^{\scriptsize \pm.602}$ & $12.7^{\scriptsize \pm.112}$ & $0.89^{\scriptsize \pm.008}$
& $17.4^{\scriptsize \pm.931}$ & $21.7^{\scriptsize \pm1.613}$ & $13.8^{\scriptsize \pm.034}$ & $0.89^{\scriptsize \pm.011}$
& $26.4^{\scriptsize \pm1.784}$ & $31.1^{\scriptsize \pm.814}$ & $9.2^{\scriptsize \pm.121}$ & $0.90^{\scriptsize \pm.007}$ \\
MotionGPT~\cite{jiang2023motiongpt}
& $15.6^{\scriptsize \pm1.321}$ & $19.3^{\scriptsize \pm.792}$ & $17.2^{\scriptsize \pm.141}$ & $0.90^{\scriptsize \pm.007}$
& $22.6^{\scriptsize \pm1.108}$ & $27.4^{\scriptsize \pm1.504}$ & $9.4^{\scriptsize \pm.159}$ & $0.91^{\scriptsize \pm.010}$
& $24.1^{\scriptsize \pm.615}$ & $29.2^{\scriptsize \pm1.953}$ & $9.7^{\scriptsize \pm.038}$ & $0.91^{\scriptsize \pm.008}$ \\
MG-MotionLLM~\cite{wu2025mgmotionllmunifiedframeworkmotion}
& $18.0^{\scriptsize \pm.841}$ & $22.9^{\scriptsize \pm1.373}$ & $12.4^{\scriptsize \pm.184}$ & $0.91^{\scriptsize \pm.007}$
& $19.3^{\scriptsize \pm1.632}$ & $24.4^{\scriptsize \pm.916}$ & $12.9^{\scriptsize \pm.039}$ & $0.90^{\scriptsize \pm.006}$
& $28.1^{\scriptsize \pm.587}$ & $33.4^{\scriptsize \pm1.822}$ & $8.7^{\scriptsize \pm.131}$ & $0.91^{\scriptsize \pm.004}$ \\
MotionLLM~\cite{chen2024motionllmunderstandinghumanbehaviors}
& $20.1^{\scriptsize \pm1.291}$ & $24.8^{\scriptsize \pm1.503}$ & $11.9^{\scriptsize \pm.145}$ & $0.90^{\scriptsize \pm.009}$
& $18.4^{\scriptsize \pm.954}$ & $23.1^{\scriptsize \pm.603}$ & $13.2^{\scriptsize \pm.151}$ & $0.91^{\scriptsize \pm.008}$
& $30.2^{\scriptsize \pm1.764}$ & $35.4^{\scriptsize \pm1.112}$ & $8.1^{\scriptsize \pm.056}$ & $0.91^{\scriptsize \pm.006}$ \\
MotionGPT3~\cite{zhu2025motiongpt3humanmotionsecond}
& $18.3^{\scriptsize \pm.713}$ & $23.1^{\scriptsize \pm1.971}$ & $12.1^{\scriptsize \pm.140}$ & $0.92^{\scriptsize \pm.004}$
& $26.2^{\scriptsize \pm1.624}$ & $31.2^{\scriptsize \pm.802}$ & $8.4^{\scriptsize \pm.085}$ & $0.91^{\scriptsize \pm.007}$
& $27.9^{\scriptsize \pm.599}$ & $33.8^{\scriptsize \pm1.237}$ & $8.9^{\scriptsize \pm.107}$ & $0.92^{\scriptsize \pm.003}$ \\
\midrule
\textbf{OmniMoGen} 
& $39.1^{\scriptsize \pm.469}$ & $50.6^{\scriptsize \pm.546}$ & $5.7^{\scriptsize \pm.076}$ & $0.95^{\scriptsize \pm.008}$
& $30.8^{\scriptsize \pm.472}$ & $43.4^{\scriptsize \pm.562}$ & $6.4^{\scriptsize \pm.034}$ & $0.96^{\scriptsize \pm.005}$
& $42.5^{\scriptsize \pm.471}$ & $48.0^{\scriptsize \pm.549}$ & $5.8^{\scriptsize \pm.198}$ & $0.94^{\scriptsize \pm.009}$ \\
\textbf{OmniMoGen-Think} 
& $\mathbf{42.9}^{\scriptsize \pm.493}$ & $\mathbf{55.4}^{\scriptsize \pm.590}$ & $\mathbf{5.3}^{\scriptsize \pm.057}$ & $\mathbf{0.97}^{\scriptsize \pm.004}$
& $\mathbf{34.7}^{\scriptsize \pm.497}$ & $\mathbf{49.3}^{\scriptsize \pm.595}$ & $\mathbf{5.8}^{\scriptsize \pm.098}$ & $\mathbf{0.97}^{\scriptsize \pm.007}$
& $\mathbf{44.3}^{\scriptsize \pm.495}$ & $\mathbf{51.9}^{\scriptsize \pm.592}$ & $\mathbf{5.5}^{\scriptsize \pm.062}$ & $\mathbf{0.96}^{\scriptsize \pm.008}$ \\
\bottomrule
\end{tabular}
}

\label{tab:anycontext}
\end{table*}

We compare \mbox{OmniMoGen} and \mbox{OmniMoGen-Think} with existing methods that support interleaved text–motion inputs on AnyContext.
From Table~\ref{tab:anycontext}, we summarize two key observations about current baselines:
\textbf{1)~Existing baselines struggle with interleaved instructions}. Across Style, Trajectory, and Speed types, their R@1 remains consistently low (often below 30), indicating that current methods have difficulty understanding free-form interleaved inputs and generating motions that correctly match semantics. 
\textbf{2)~They show limited physical plausibility}. 
Their Physical metrics are uniformly below 0.91, showing a clear gap from the Real Data of 0.98. This suggests notable instability in foot contact when handling interleaved generation.

Compared to existing methods, OmniMoGen achieves the following advantages: 
\textbf{1)~Higher semantic accuracy and physical plausibility}.
\mbox{OmniMoGen} achieves an average R@1 of 37.5 across the three task types, outperforming the second-best baseline by 15.13. 
Its Physical metric reaches 0.97, clearly higher than baselines and approaching the Real Data of 0.98.
\textbf{2)~Further performance improvement through reflection}.
On average across Style, Trajectory, and Speed tasks, \mbox{OmniMoGen-Think} improves R@1 by 3.17 compared to \mbox{OmniMoGen}, and increases Physical by 0.02.

\subsection{Emerging Capabilities}
\label{sec4:emerge}

\begin{figure}
    \centering
    \includegraphics[width=\linewidth]{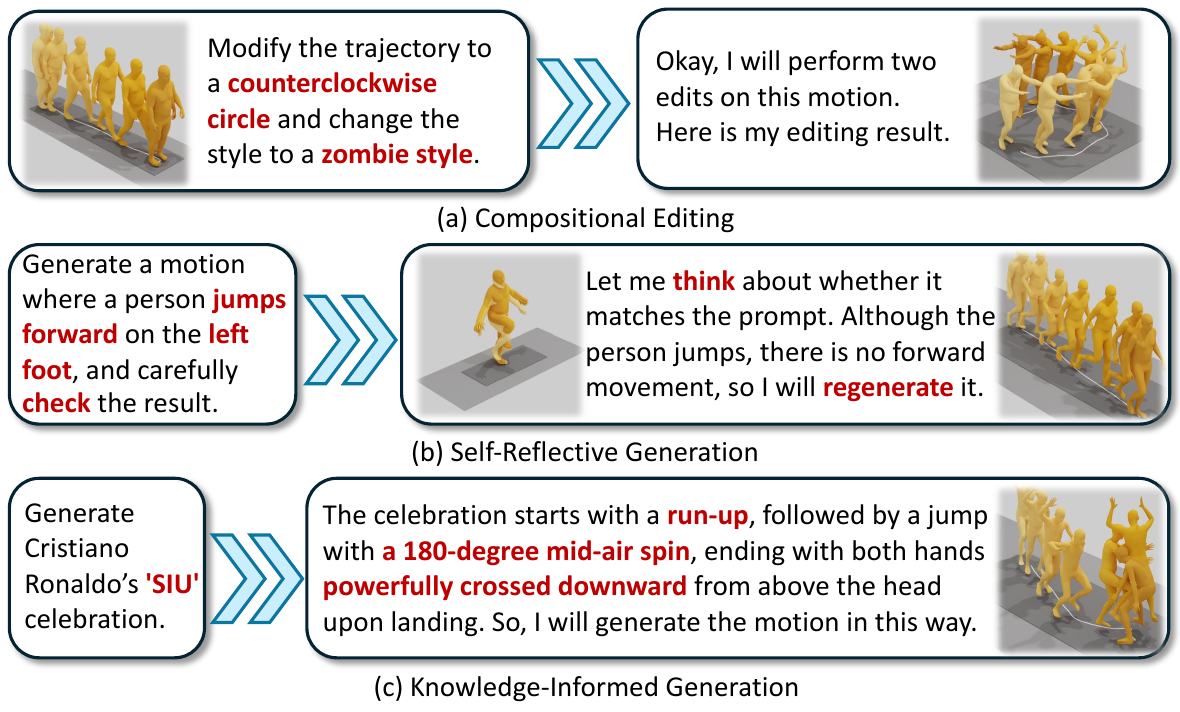}
    \caption{Examples of emerging capabilities for OmniMoGen.}
    \label{fig:emerging}
\end{figure}

Beyond specific motion generation tasks, OmniMoGen also shows emerging capabilities, as shown in Figure~\ref{fig:emerging}. 

\noindent\textbf{Compositional Editing.}
In real-world applications, user requirements often involve combinations of various adjustments.
OmniMoGen can perform compositional editing, where multiple sequential edits can be accomplished through a single instruction. 
For example, instead of two independent edits, the model can directly execute a combined instruction like ``modify the trajectory to a counterclockwise circle and change the style to a zombie style,'' as shown in Figure~\ref{fig:emerging}(a).

\noindent\textbf{Self-Reflective Generation.}
OmniMoGen also exhibits a self-reflective behavior during motion generation. 
We observe that the model can assess whether the generated motion aligns with the instruction when it is asked to carefully check the result.
If a mismatch occurs, it tends to regenerate the motion for correction, as shown in Figure~\ref{fig:emerging}(b).

\noindent\textbf{Knowledge-Informed Generation.}
Benefiting from two-stage training with interleaved text–motion instructions, the model acquires strong reasoning capabilities and broad world knowledge.
As shown in Figure~\ref{fig:emerging}(c), OmniMoGen infers a description of the `SIU' celebration and then generates this previously unseen motion following the motion description.

\subsection{In-Depth Analysis}
\label{sec4:analysis}

\noindent
\textbf{Analysis of Test-time Scaling.}
Following the idea of test-time scaling in LLMs~\cite{guo2025deepseek}, we let OmniMoGen perform additional rounds of reflection.
We set the maximum number of reflection rounds to 1, 2, and 3, respectively.
As shown in Table~\ref{tab:ablation}(A), increasing the number of reflection rounds leads to more accurate generation, with the AvgR on MotionFix~\cite{athanasiou2024motionfixtextdriven3dhuman} decreasing from 2.47 to 2.31 and the AvgR on AnyContext decreasing from 6.1 to 5.4. 
In addition, we observe that reflection also improves the Physical metric on AnyContext from 0.95 to 0.97.
We provide more details in Appendix~\textcolor{red}{D.1}.

\noindent
\textbf{Effect of Training Strategy.} 
To assess the importance of each stage in our two-stage training, we remove SFT or GRPO individually.
As shown in Table~\ref{tab:ablation}(B), removing SFT causes a large drop in retrieval accuracy on both datasets, indicating that supervised multi-task learning is the key to establishing basic motion generation ability.
Removing GRPO also reduces performance, especially on AnyContext, where AvgR rises from 6.1 to 8.0 and Physical decreases from 0.95 to 0.91, demonstrating GRPO effectively improves semantic correctness and physical plausibility.

\noindent
\textbf{Effect of Reward in GRPO.} 
To analyze the contribution of rewards, we remove them during GRPO.
As shown in Table~\ref{tab:ablation}(C), removing the semantic correctness reward decreases MotionFix~\cite{athanasiou2024motionfixtextdriven3dhuman} R@1 from 68.33 to 63.62 and increases AvgR from 2.47 to 2.69.
Removing the physical plausibility reward reduces the Physical metric from 0.95 to 0.92 on AnyContext.
This indicates that the two rewards complement each other in improving the semantic accuracy and physical plausibility of OmniMoGen.

\noindent
\textbf{Synergy of Multi-tasks in SFT.} 
To understand how subtasks contribute during SFT, we remove them individually.
As shown in Table~\ref{tab:ablation}(D), removing motion editing lowers MotionFix~\cite{athanasiou2024motionfixtextdriven3dhuman} R@1 from 68.33 to 56.25, and removing multi-turn editing decreases it to 59.07, indicating their importance for handling more complex edit instructions.

\noindent
\textbf{Effect of Model Components.}
We first evaluate different language backbones while fixing RVQ-VAE as the motion tokenizer.
As shown in Table~\ref{tab:ablation}(E), model performance generally improves with larger backbones: R@1 increases from 48.82 with T5-Small~\cite{raffel2020exploring} to 68.33 with Gemma2-2B~\cite{gemmateam2024gemma2improvingopen}.
Gemma2-9B shows a slight drop, which we attribute to overfitting caused by the limited size of our training corpus.
Based on this trade-off between performance and stability, we adopt Gemma2-2B as the language backbone for \mbox{OmniMoGen}.
After determining the backbone, we further analyze the contribution of the motion tokenizer.
As shown in Table~\ref{tab:ablation}(F), RVQ-VAE consistently outperforms VQ-VAE, improving MotionFix~\cite{athanasiou2024motionfixtextdriven3dhuman} R@1 from 64.91 to 68.33, indicating that the residual-quantized representation is better suited for high-precision motion generation.

\begin{table}[t!]
\centering

\caption{
Comprehensive ablation study of OmniMoGen. 
We report Edited-to-Target R@1 and AvgR on MotionFix, and R@1, AvgR, and Physical across all task types in AnyContext.
}
\resizebox{\linewidth}{!}{
\begin{tabular}{lccccc}
\toprule
\multirow{2}{*}{\textbf{Methods}}
& \multicolumn{2}{c}{\textbf{MotionFix}} 
& \multicolumn{3}{c}{\textbf{AnyContext}} \\
\cmidrule(lr){2-3} \cmidrule(lr){4-6}
 & R@1$\uparrow$ & AvgR$\downarrow$
 & R@1$\uparrow$ & AvgR$\downarrow$ & Physical$\uparrow$ \\
\midrule

\textbf{OmniMoGen} 
& 68.33 & 2.47 & 36.7 & 6.1 & 0.95 \\

\midrule
\multicolumn{6}{l}{\textbf{(A) Max Reflection Rounds}} \\
w/ 1 Round     & 69.12 & 2.42 & 39.5 & 5.8 & 0.96 \\
w/ 2 Rounds    & 70.39 & 2.37 & 40.9 & 5.6 & 0.96 \\
w/ 3 Rounds    & 71.59 & 2.31 & 41.2 & 5.4 & 0.97 \\

\midrule
\multicolumn{6}{l}{\textbf{(B) Training Strategy}} \\
w/o SFT  & 42.09 & 5.85 & 19.3 & 12.4 & 0.94 \\
w/o GRPO   & 61.83 & 2.84 & 30.8 & 8.0 & 0.91 \\

\midrule
\multicolumn{6}{l}{\textbf{(C) GRPO Reward}} \\
w/o Semantic Correctness Reward  & 63.62 & 2.69 & 30.5 & 7.9 & 0.94 \\
w/o Physical Plausibility Reward   & 67.28 & 2.52 & 34.9 & 7.0 & 0.92 \\

\midrule
\multicolumn{6}{l}{\textbf{(D) Multi-Task in SFT}} \\
w/o Motion Editing          & 56.25 & 3.23 & 34.2 & 7.2 & 0.94 \\
w/o Multi-Turn Editing          & 59.07 & 2.92 & 35.4 & 6.7 & 0.95 \\
w/o In-Context Generation       & 65.59 & 2.67 & 30.2 & 8.0 & 0.95 \\
w/o Reflection          & 67.46 & 2.54 & 34.3 & 7.1 & 0.94 \\

\midrule
\multicolumn{6}{l}{\textbf{(E) Language Backbone}} \\
T5-Small~(60M)     & 48.82 & 4.73 & 25.4 & 10.3 & 0.87 \\
T5-Base~(220M)      & 56.26 & 3.27 & 32.0 & 7.5 & 0.90 \\
T5-Large~(770M)      & 60.74 & 2.89 & 34.1 & 7.3 & 0.91 \\
Gemma2-2B    & 68.33 & 2.47 & 36.7 & 6.1 & 0.95 \\
Gemma2-9B    & 67.17 & 2.58 & 36.8 & 6.3 & 0.95 \\

\midrule
\multicolumn{6}{l}{\textbf{(F) Motion Tokenizer}} \\
VQ-VAE          & 64.91 & 2.66 & 30.9 & 8.2 & 0.93 \\
RVQ-VAE         & 68.33 & 2.47 & 36.7 & 6.1 & 0.95 \\

\bottomrule
\end{tabular}
}

\label{tab:ablation}
\end{table}

\section{Conclusion}
\label{sec:conclusion}
In this work, we introduce \textbf{OmniMoGen}, a unified framework that accomplishes diverse human motion generation tasks through free-form instruction following. To support such unification, we build \textbf{X2Mo}, the first large-scale interleaved text–motion dataset, and introduce \textbf{AnyContext} for evaluating interleaved motion generation. Through two-stage training on X2Mo, OmniMoGen achieves state-of-the-art performance on AnyContext, HumanML3D, and MotionFix. 
Additionally, OmniMoGen exhibits emerging capabilities such as knowledge-informed generation, marking an important step toward the next generation of motion generation.

{
    \small
    \bibliographystyle{ieeenat_fullname}
    \bibliography{main}
}

\addtocontents{toc}{\protect\setcounter{tocdepth}{3}}
\clearpage
\maketitlesupplementary

\appendix
\renewcommand{\contentsname}{\textbf{Table of Contents in Appendix}} 
\renewcommand{\thesection}{\Alph{section}}
\renewcommand{\thesubsection}{\thesection.\arabic{subsection}}
\renewcommand{\cftsecleader}{\cftdotfill{\cftdotsep}} 
\hypersetup{linkcolor=black}
\tableofcontents
\hypersetup{linkcolor=blue}

\section{Dataset Construction}

\subsection{Statistics of X2Mo}
X2Mo contains 137K interleaved text–motion instructions covering four major task types: in-context generation, motion editing, multi-turn editing, and reflection.
All motions are derived from AMASS motion capture sequences and segmented into units with a duration of up to 10 seconds at 20 fps, aligned with HumanML3D~\cite{guo2022generating} formatting for fair comparison.
We remove all sequences appearing in the test sets of HumanML3D and MotionFix~\cite{athanasiou2024motionfixtextdriven3dhuman} to avoid data leakage. The dataset contains a diverse set of motions such as walking, running, jumping, fighting, athletic skills, and daily activities, each annotated with structured attributes including action type, style, trajectory, duration, and speed. 

\begin{figure}
    \centering
    \includegraphics[width=\linewidth]{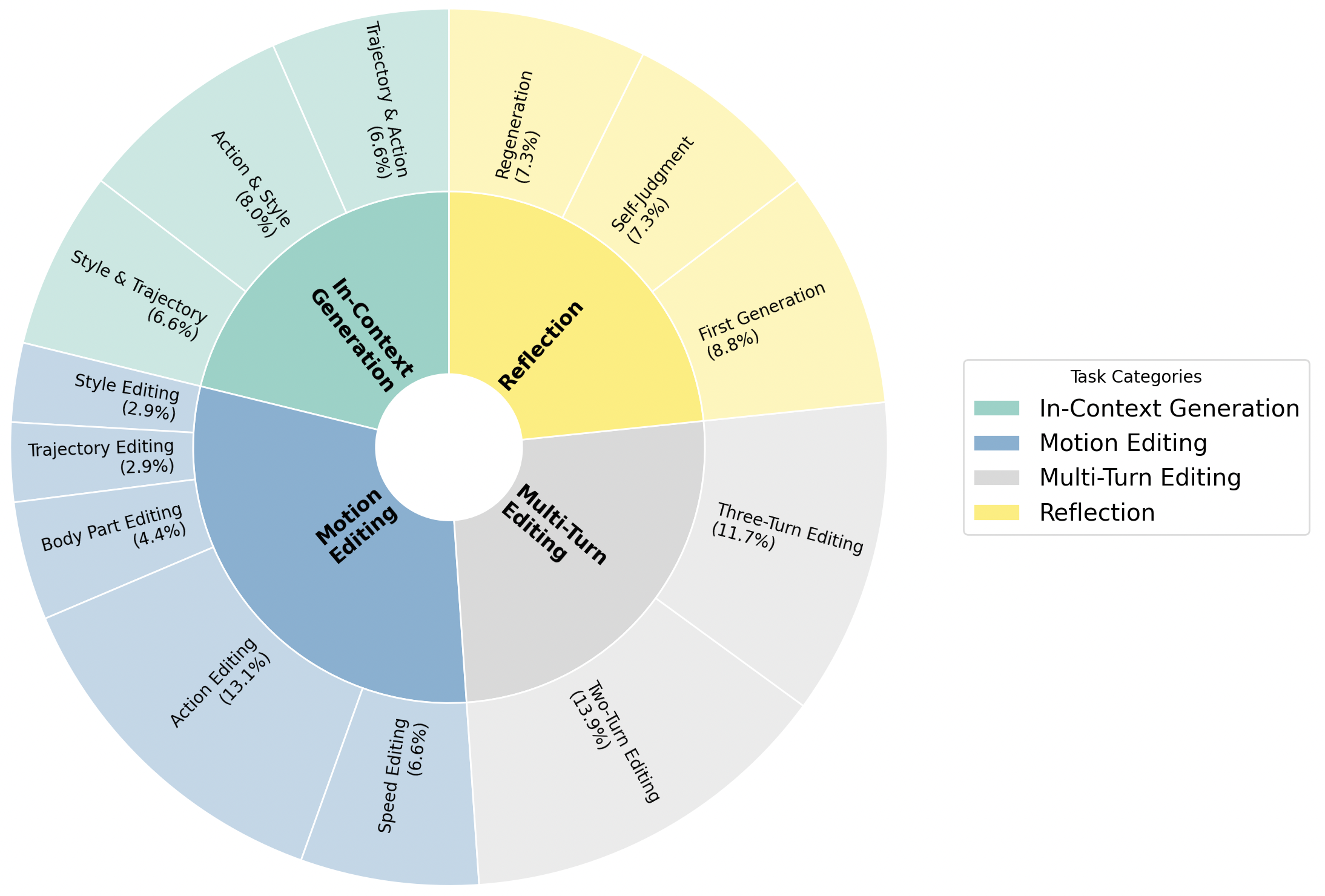}
    \caption{The dataset is organized into four high-level task categories and their fine-grained subtasks. The inner ring shows the proportions of each category, while the outer ring presents the distribution of corresponding subtasks.}
    \label{fig:statistics}
\end{figure}

\noindent
\textbf{Task Distribution.} X2Mo covers four types of interleaved text–motion instructions. 1)~In-context generation focuses on composing action, style, and trajectory elements from multiple reference motions to enable compositional generation. It contains approximately 29K samples. 2)~Motion editing provides paired source–target motions that differ in action primitives, spatial paths, or locomotion constraints, allowing fine-grained modification based on textual instructions. It contains approximately 41K samples. 3)~Multi-turn editing simulates iterative refinement by forming sequential text–motion interaction chains. It contains approximately 35K samples. 4)~Reflection includes both aligned and misaligned motion–caption pairs to support introspective reasoning and self-correction. It contains approximately 32K samples. We provide a detailed task distribution in Figure~\ref{fig:statistics}.

\subsection{Data Pipeline}

We construct X2Mo using an automated pipeline that transforms raw motion capture sequences into interleaved text–motion instructions suitable for unified motion generation. The pipeline comprises three stages: motion preprocessing, motion graph construction, and instruction interleaving from the graph.

\subsubsection{Motion Preprocessing}

The AMASS~\cite{mahmood2019amassarchivemotioncapture} dataset serves as the primary source of high-quality 3D human motion used in constructing our dataset. 
We denote a raw motion sequence as $\mathbf{M} = \{\mathbf{p}_t\}_{t=1}^{T}$, where $\mathbf{p}_t \in \mathbb{R}^{J \times d}$ represents $J$ joint parameters in $d$-dimensional space at timestamp $t$.
To ensure consistency with HumanML3D, all motion sequences are downsampled to a unified frame rate of $20$ fps. 
Motion sequences longer than 10 seconds are further segmented into action-consistent units based on BABEL~\cite{punnakkal2021babelbodiesactionbehavior} frame annotations. 
Formally, we define the segmentation function
\[
\text{Seg}(\mathbf{M}) = \{\mathbf{M}^{(1)}, \mathbf{M}^{(2)}, \dots, \mathbf{M}^{(K)}\},
\]
where each segment maintains consistent action semantics throughout its duration. 
To prevent information leakage in evaluation, we remove all motion segments appearing in test splits of HumanML3D and MotionFix. 

\subsubsection{Motion Graph Construction}
We adopt the motion encoder introduced in TMR to map each motion segment into a shared embedding space. Let $\mathbf{M}^{(i)}$ denote the $i$-th motion segment and $f(\cdot)$ be the encoder; its embedding is given by
\[
\mathbf{z}_i = f(\mathbf{M}^{(i)}).
\]
We compute pairwise similarity across all segment embeddings using cosine similarity
\[
\text{sim}(\mathbf{z}_i, \mathbf{z}_j) = \frac{\langle \mathbf{z}_i, \mathbf{z}_j \rangle}{\|\mathbf{z}_i\| \, \|\mathbf{z}_j\|}.
\]
A directed edge from segment $i$ to segment $j$ is created when their similarity surpasses a threshold $\tau = 0.9$:
\[
\text{sim}(\mathbf{z}_i, \mathbf{z}_j) > \tau.
\]
To avoid cycles during graph construction, we enforce a directional constraint such that the source node index is always smaller than the target index $(i < j)$, ensuring all edges form a directed acyclic structure.
For each motion segment, we further render it into a short video and employ the state-of-the-art video understanding model \mbox{Gemini-2.5-Pro} to annotate fine-grained action information. Each annotated action is represented using a structured tuple
\[
a = (\text{type}, \text{body-part}, \text{style}, \text{duration}, \text{trajectory}),
\]
forming an action list for each segment. If two connected segments produce identical action lists, the corresponding edge is discarded as it does not introduce a meaningful transformation.
After filtering, all remaining edges collectively form a directed acyclic motion graph, referred to as the motion graph.

\subsubsection{Interleaving from Motion Graph}
Given the constructed motion graph $\mathcal{G} = (\mathcal{V}, \mathcal{E})$, where each node corresponds to a semantically coherent motion segment and each directed edge $(i \rightarrow j) \in \mathcal{E}$ denotes a meaningful transformation in motion attributes, we convert graph structures into four types of interleaved text--motion instructions. This process transforms the structural properties of the motion graph into natural language prompts that guide the model to perform reasoning, editing, composition, and self-verification over motion sequences. Collectively, these instructions form the foundation for unified multi-task learning.

\textit{(1) In-context generation.}
To construct in-context instructions, we search for subgraphs demonstrating convergent topology, where multiple source nodes point to a shared target node:
\[
(i_1 \rightarrow k), (i_2 \rightarrow k), \ldots, (i_m \rightarrow k).
\]
Such structures imply that the target motion can be explained as a composition of distinct motion elements extracted from the source motions. Let $A_x$ denote the action list of node $x$. If the action list of the target motion satisfies:
\[
A_k \subseteq \bigcup_{\ell=1}^{m} A_{i_\ell},
\]
then the model can theoretically reassemble the target motion by retrieving different attributes (e.g., action type, style, or trajectory) from each source motion. Based on this observation, we construct natural language prompts such as ``concatenate walking from motion A with turning from motion B.'' These samples encourage the model to perform compositional inference, mimicking in-context learning behaviors commonly studied in LLMs, but within the motion domain.

\textit{(2) Motion editing.}
For each directed edge $(i \rightarrow j)$, we regard node $i$ as the source motion and node $j$ as the modified version of that motion. We compute the semantic difference between their action lists:
\[
\Delta A = A_j \setminus A_i,
\]
which represents the behavioral modification introduced in the target motion. If $\Delta A$ contains additional elements such as a kick or a trajectory shift, we generate an editing instruction describing the incremental change, e.g., ``add a kicking at the end'' or ``turn left while maintaining the same pace.'' Unlike in-context generation, which merges attributes from multiple motions, motion editing isolates a single directional transformation and formalizes it as a localized instruction. This enables the model to learn how motion transitions correspond to concise textual commands without altering unrelated motion attributes.

\textit{(3) Multi-turn editing.}
To simulate realistic interaction scenarios involving iterative refinement, we sample continuous directed paths from the motion graph:
\[
i_1 \rightarrow i_2 \rightarrow \cdots \rightarrow i_T.
\]
Each node represents a successive motion state, and each edge encodes a minimal semantic edit from one state to the next. Starting from $i_1$ as the initial motion, we generate a sequence of editing instructions that iteratively transform $\mathbf{M}^{(i_t)}$ into $\mathbf{M}^{(i_{t+1})}$. Each instruction is derived using the same action-difference computation as motion editing but applied across multiple steps. This yields multi-step editing trajectories such as:
\[
\{\text{walk forward}\} \rightarrow \{\text{walk faster}\} \rightarrow \{\text{run in a circle}\}.
\]
These samples encourage the model to maintain temporal coherence across successive edits, ensuring that each refinement builds upon the previous motion rather than resetting semantics. Such iterative structures are crucial for multi-round human–AI co-creation or interactive animation editing.

\textit{(4) Reflection.}
Beyond generative supervision, we further construct reflection samples aimed at teaching the model to evaluate whether a generated motion aligns with a given textual description. For each edge $(i \rightarrow j)$, we take the caption associated with the target motion as the input description. We construct positive samples by pairing this caption with motion $j$, where semantics match, and negative samples by pairing the same caption with motion $i$, where motion and language are misaligned.
These contrastive pairs enable explicit supervision on judgment and evaluation rather than generation alone. We further employ MLLMs to produce structured reasoning traces explaining \emph{why} a motion matches or mismatches the caption. This transforms the task from binary discrimination into introspective reasoning, supporting behaviors such as self-correction, error diagnosis, and iterative improvement during motion generation.

Collectively, the four types of interleaved instructions enable the model to learn generation, refinement, composition, and reflection in a unified manner, establishing a general-purpose motion agent capable of free-form instruction following.

\section{Benchmark Details}
\subsection{Task Types}
AnyContext evaluates interleaved motion generation where each test sample consists of a source motion, a reference motion, and a natural language instruction. Importantly, all tasks are defined under a consistent constraint: the generated motion must preserve the core action semantics derived from the source motion, while selectively inheriting specific attributes from the reference motion. This design explicitly evaluates the capability of a model to perform targeted modification rather than unconditional generation, reflecting realistic usage scenarios where users refine existing motions under free-form multimodal instructions.
We categorize interleaved tasks into three types: style-based, trajectory-based, and speed-based, each corresponding to a distinct motion attribute extracted from the reference motion.

\noindent
\textbf{Style-based Generation.}
This task evaluates whether the model can maintain the action category of the source motion while transferring the reference motion's stylistic properties (e.g., zombie-like, energetic, cautious). Since style represents high-level behavioral characteristics independent of spatial path or action semantics, this task probes a model’s ability to identify and transfer abstract expressive attributes from motion examples.

\noindent
\textbf{Trajectory-based Generation.}
In trajectory-based generation, the action semantics from the source motion remain unchanged, but the generated motion must follow the spatial path demonstrated in the reference motion. Examples include walking along a clockwise arc, stepping diagonally forward, or moving backward while performing the same action. This task measures the model’s spatial controllability and its ability to recombine motion structure while maintaining temporal coherence.

\noindent
\textbf{Speed-based Generation.}
Speed-based generation modifies temporal execution, such as slowing down, accelerating, or matching rhythmic pacing based on the reference motion. While the action type and trajectory follow the source motion, the temporal profile should align with the reference. This task evaluates temporal alignment and fine-grained control over motion kinematics.

Across all task types, AnyContext enforces a compositional instruction-following paradigm: the source motion determines \emph{what} action is performed, while the reference motion determines \emph{how} it is performed along different semantic dimensions. This design enables a fine-grained evaluation of interleaved motion generation ability, distinguishing models that merely produce plausible motions from those capable of attribute-level controlled generation under multimodal supervision.

\subsection{Evaluation Metrics}

For AnyContext, we evaluate generated motions along two complementary dimensions: (a) semantic correctness, assessed using retrieval accuracy based on feature similarity, and (b) physical plausibility, assessed using foot–ground consistency. Both metrics measure distinct aspects of interleaved instruction following: semantic alignment with target motion attributes and biomechanical realism during execution.

\noindent
\textbf{Retrieval Accuracy.}
We follow a retrieval-based evaluation protocol. Given a generated motion sequence $\hat{M}$ conditioned on an interleaved instruction $(M_{\text{src}}, M_{\text{ref}}, T)$, we compute embedding representations using a pretrained motion-text encoder $f(\cdot)$. Let $\mathcal{G} = \{M_1, M_2, \dots, M_K\}$ be a randomly sampled gallery (typically $K=32$). We compute cosine similarity between the generated motion and all candidates:
\[
s_i = \text{sim}\left(f(\hat{M}), f(M_i)\right) = \frac{f(\hat{M})^\top f(M_i)}{\lVert f(\hat{M})\rVert_2 \lVert f(M_i)\rVert_2}.
\]

We obtain a ranking by sorting $\{s_i\}$ in descending order. Retrieval metrics are computed as follows:
\[
\text{R@k} = \frac{1}{N} \sum_{j=1}^{N} \mathbb{1}\!\left[\text{rank}(\hat{M}_j, M_{j}^{\text{target}}) \le k\right],
\]
where $\text{rank}(\cdot)$ is the position of the target motion in the ranked list, and $N$ is the number of test samples.

To holistically summarize ranking performance, we compute the average rank:
\[
\text{AvgR} = \frac{1}{N} \sum_{j=1}^{N} \text{rank}(\hat{M}_j, M_{j}^{\text{target}}),
\]
where lower $\text{AvgR}$ indicates better semantic alignment. Retrieval metrics capture whether the generated motion reflects attributes specified by the interleaved instruction (e.g., applying the reference motion's style while preserving the source motion's action).

\noindent
\textbf{Physical Plausibility.}
To evaluate biomechanical realism, we compute a foot contact score that penalizes sliding and floating. For each foot joint $i \in \{\text{L-ankle}, \text{L-toe}, \text{R-ankle}, \text{R-toe}\}$ and frame $t$, let $z^t_i$ denote foot-ground height and $v^t_i$ horizontal velocity. We define:
\[
d^t_i = \max(|z^t_i| - \tau_h, 0),
u^t_i = \max(\|v^t_i\|_2 - \tau_v, 0),
\]
where $\tau_h$ and $\tau_v$ are thresholds controlling when the motion is considered non-contact ($\tau_h = 0.05$m, $\tau_v = 0.075$m/s). The per-joint contact score is defined as:
\[
s^t_i = \exp(-d^t_i) \cdot \exp(-u^t_i).
\]

Averaging across time and joints yields:
\[
S_{\text{phy}} = \frac{1}{T \cdot 4}\sum_{t=1}^{T}\sum_{i=1}^{4} s^t_i.
\]

Higher $S_{\text{phy}}$ indicates stable grounding and reduced artifacts such as skating or mid-air dragging. This metric complements retrieval accuracy: motions can be semantically aligned while physically implausible, and vice versa.

Together, these two metrics reflect the core challenge of AnyContext: generating motions that not only obey interleaved semantic signals but also remain biomechanically valid during execution.

\section{Interleaved Instructions in X2Mo}

In this section, we present the example templates used to construct the interleaved text-motion instructions in X2Mo. These templates define how each fine-grained task is represented in natural language, serving as the core interface between textual descriptions and motion sequences. 

\subsection{In-Context Generation}

This category focuses on synthesizing new motions by reusing semantic elements from existing motions. Instead of directly executing a single reference example, the instructions specify which aspects of different motions, such as action, style, or trajectory, should be extracted and recombined. These interleaved instructions emphasize compositional reuse rather than imitation, allowing the model to generalize motion attributes across diverse contexts. We provide some example instructions in Table~\ref{tab:sup1}.

\begin{table*}[t]
\centering
\caption{Prompt templates for fine-grained tasks within in-context generation.}
\label{tab:sup1}
\resizebox{\linewidth}{!}{
\begin{tabular}{p{3cm} p{3.5cm} p{10cm} p{2cm}}
\toprule
\textbf{Primary Task} & \textbf{Fine-Grained} & \textbf{Example Instructions} & \textbf{Output} \\
\midrule

\multirow{3}{*}{In-Context Generation} 
& Action + Style &
\begin{itemize}
    \item Perform the action from \textless Motion\textgreater{} in the style of \textless Motion\textgreater{}.
    \item Use the action from \textless Motion\textgreater{} and apply the style of \textless Motion\textgreater{}.
    \item Reproduce the same action as \textless Motion\textgreater{} but follow the style of \textless Motion\textgreater{}.
\end{itemize}
& motion \\
\cmidrule(lr){2-4}

& Style + Trajectory &
\begin{itemize}
    \item Follow the trajectory of \textless Motion\textgreater{} while using the style of \textless Motion\textgreater{}.
    \item Move along the path of \textless Motion\textgreater{} while expressing the style of \textless Motion\textgreater{}.
    \item Use the spatial route from \textless Motion\textgreater{} combined with the style of \textless Motion\textgreater{}.
\end{itemize}
& motion \\
\cmidrule(lr){2-4}

& Action + Trajectory &
\begin{itemize}
    \item Perform the action from \textless Motion\textgreater{} while following the trajectory of \textless Motion\textgreater{}.
    \item Execute the action of \textless Motion\textgreater{} using the path in \textless Motion\textgreater{}.
    \item Use the route from \textless Motion\textgreater{} but keep the action consistent with \textless Motion\textgreater{}.
\end{itemize}
& motion \\

\bottomrule
\end{tabular}
}
\end{table*}

\subsection{Motion Editing}

This category specifies how a given motion should be modified along a single dimension while preserving its original semantic identity. Editing instructions target specific attributes such as action, style, trajectory, speed, or body parts, enabling localized changes without altering unrelated motion components. This setup allows models to learn fine-grained and controlled modification behavior rather than full regeneration. We provide some example instructions in Table~\ref{tab:sup2}.

\begin{table*}[t]
\centering
\caption{Prompt templates for fine-grained tasks within motion editing.}
\label{tab:sup2}
\resizebox{\linewidth}{!}{
\begin{tabular}{p{3cm} p{3.5cm} p{10cm} p{2cm}}
\toprule
\textbf{Primary Task} & \textbf{Fine-Grained} & \textbf{Example Instructions} & \textbf{Output} \\
\midrule

\multirow{5}{*}{Motion Editing} 
& Action Editing &
\begin{itemize}
    \item Replace the final segment with a jumping action in \textless Motion\textgreater{}.
    \item Insert a kicking action at the end of \textless Motion\textgreater{}.
    \item Change the action in \textless Motion\textgreater{} from walking to crawling.
\end{itemize}
& motion \\
\cmidrule(lr){2-4}

& Style Editing &
\begin{itemize}
    \item Change the style of \textless Motion\textgreater{} to energetic.
    \item Modify \textless Motion\textgreater{} to express a cautious style.
    \item Convert the style of \textless Motion\textgreater{} to relaxed.
\end{itemize}
& motion \\
\cmidrule(lr){2-4}

& Trajectory Editing &
\begin{itemize}
    \item Change the trajectory of \textless Motion\textgreater{} to a clockwise circular path.
    \item Make \textless Motion\textgreater{} move forward instead of staying in place.
\end{itemize}
& motion \\
\cmidrule(lr){2-4}

& Speed Editing &
\begin{itemize}
    \item Slow down \textless Motion\textgreater{}.
    \item Increase the speed of \textless Motion\textgreater{}.
\end{itemize}
& motion \\
\cmidrule(lr){2-4}

& Body Part Editing &
\begin{itemize}
    \item Raise both arms in \textless Motion\textgreater{}.
    \item Add a pointing gesture with the right hand in \textless Motion\textgreater{}.
    \item Make the head face forward in \textless Motion\textgreater{}.
\end{itemize}
& motion \\

\bottomrule
\end{tabular}
}
\end{table*}

\subsection{Multi-Turn Editing}

Unlike single-turn editing, multi-turn instructions simulate an interactive refinement process in which a motion is adjusted across multiple successive steps. Each turn introduces incremental modifications, and subsequent edits operate on the previously updated result. The interleaved structure models real-world iterative workflows, enabling the model to track state evolution and accumulate changes over time. We provide some example instructions in Table~\ref{tab:sup3}.

\begin{table*}[t]
\centering
\caption{Prompt templates for fine-grained tasks within multi-turn editing.}
\label{tab:sup3}
\resizebox{\linewidth}{!}{
\begin{tabular}{p{3cm} p{3.5cm} p{10cm} p{2cm}}
\toprule
\textbf{Primary Task} & \textbf{Fine-Grained} & \textbf{Example Instructions} & \textbf{Output} \\
\midrule

\multirow{2}{*}{Multi-Turn Editing} 
& Two-Turn Editing &
\begin{itemize}
    \item (1) Slow down \textless Motion\textgreater{}. (2) Turn its movement direction to the right.
    \item (1) Convert the style of \textless Motion\textgreater{} to energetic. (2) Add a jump at the end.
\end{itemize}
& motion \\
\cmidrule(lr){2-4}

& Three-Turn Editing &
\begin{itemize}
    \item (1) Slow down \textless Motion\textgreater{}. (2) Add arm-swinging action. (3) Modify the moving trajectory to a circular path.
    \item (1) Lower the upper-body posture of \textless Motion\textgreater{}. (2) Increase its speed. (3) Add a turn at the end.
\end{itemize}
& motion \\

\bottomrule
\end{tabular}
}
\end{table*}

\subsection{Reflection}

Reflection instructions incorporate a self-monitoring process by first generating a motion based on a caption, then evaluating how well the motion matches the caption, and finally regenerating a motion based on the judgement. This chain of interleaved instructions encourages introspective reasoning, allowing the model to diagnose semantic mismatches and refine outputs based on explicit feedback. Unlike conventional supervised learning, reflection introduces a self-corrective loop that improves motion quality during generation. We provide some example instructions in Table~\ref{tab:sup4}.

\begin{table*}[t]
\centering
\caption{Prompt templates for fine-grained tasks within reflection.}
\label{tab:sup4}
\resizebox{\linewidth}{!}{
\begin{tabular}{p{3cm} p{3.5cm} p{10.5cm} p{2cm}}
\toprule
\textbf{Primary Task} & \textbf{Fine-Grained} & \textbf{Example Instructions} & \textbf{Output} \\
\midrule

\multirow{3}{*}{Reflection} 

& First Generation &
\begin{itemize}
    \item Generate a motion following \textless Caption\textgreater{}.
    \item Create a motion that corresponds to \textless Caption\textgreater{}.
    \item Produce a motion based on \textless Caption\textgreater{}.
\end{itemize}
& motion \\
\cmidrule(lr){2-4}

& Self-Judgement &
\begin{itemize}
    \item Generate a motion following \textless Caption\textgreater{}. \textless Motion\textgreater{}. Whether the motion matches the caption?
    \item \textless Caption\textgreater{}. \textless Motion\textgreater{}. Determine whether the motion aligns with the caption.
    \item \textless Caption\textgreater{}. \textless Motion\textgreater{}. Does the motion match the caption?
\end{itemize}
& text \\
\cmidrule(lr){2-4}

& Regeneration &
\begin{itemize}
    \item Generate a motion following \textless Caption\textgreater{}. \textless Motion\textgreater{}. Whether the motion matches the caption? No, they do not match because \textless Reason\textgreater{}. I will regenerate a motion.
    \item \textless Caption\textgreater{}. \textless Motion\textgreater{}. Whether the motion matches the caption? No, they do not match because \textless Reason\textgreater{}. I will generate a new motion.
    \item Generate a motion following \textless Caption\textgreater{}. \textless Motion\textgreater{}. Whether the motion matches the caption? No, they do not match because \textless Reason\textgreater{}. I will correct it and regenerate the motion.
\end{itemize}
& motion \\

\bottomrule
\end{tabular}
}
\end{table*}

\section{Details of In-Depth Analysis}

\subsection{Analysis of Test-time Scaling}

\begin{figure}[t]
    \centering
    \begin{subfigure}{0.48\linewidth}
        \centering
        \includegraphics[width=\linewidth]{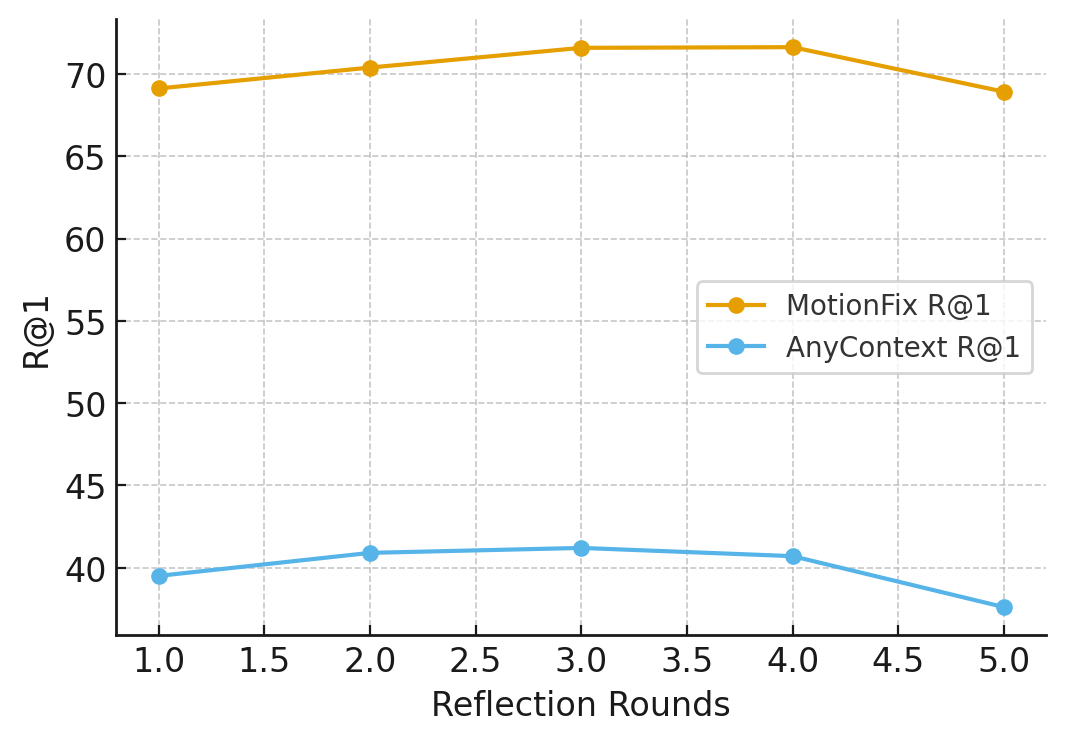}
        \caption{R@1 performance}
        \label{fig:reflection-r1}
    \end{subfigure}
    \hfill
    \begin{subfigure}{0.48\linewidth}
        \centering
        \includegraphics[width=\linewidth]{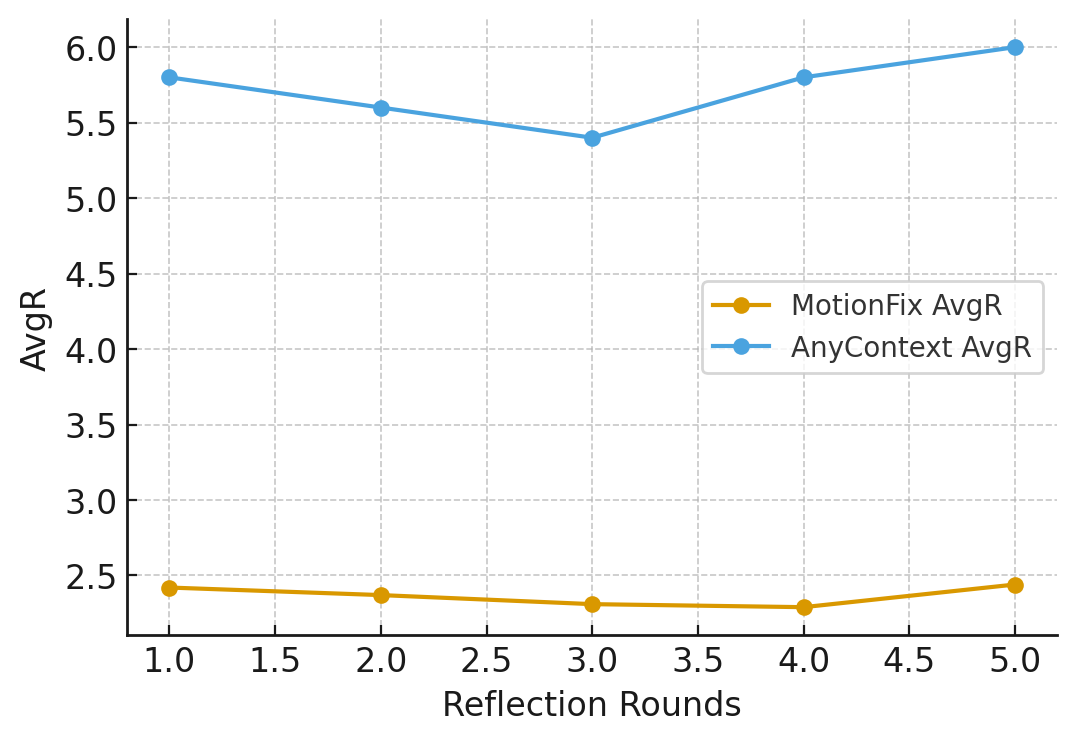}
        \caption{AvgR performance}
        \label{fig:reflection-avgr}
    \end{subfigure}
    \caption{Effect of reflection rounds on model performance across MotionFix and AnyContext.}
    \label{fig:reflection-paired}
\end{figure}

To further investigate the effect of test-time scaling in reflection-based generation, we extend the maximum number of reflection rounds beyond the setting used in the main paper. In the main experiments, OmniMoGen is evaluated under up to three rounds of reflection, demonstrating consistent performance improvements (e.g., decreasing the AvgR on AnyContext from 6.1 to 5.4, and increasing the Physical metric from 0.95 to 0.97). In this section, we explore a larger reflection budget and conduct additional experiments with up to five reflection rounds.

As shown in Figure~\ref{fig:reflection-paired}, we evaluate performance on both MotionFix and AnyContext, following the same metrics used in the main paper. Across reflection rounds 0–5, we observe a clear trend: reflection initially improves semantic alignment and physical plausibility, but only up to a certain point. Specifically, increasing rounds from 1 to 3 steadily improves retrieval performance, while 3 and 4 rounds yield comparable results with marginal differences. However, extending to 5 rounds leads to performance degradation. We attribute this degradation to limitations in the model’s effective context window.

\section{Evaluation Details}

\subsection{Baselines}
With the rapid development of MLLMs~\cite{li2023variational,li2023fine,li2025structure,pan2025generative,pan2024auto,panjanus,pan2025focusdiff,pan2025wiseedit,pan2024towards,Yu_2025_CVPR,gao2025arcadiafulllifecycleframeworkembodied,gao2024generalistvirtualagentssurvey}, a wide range of methods for human motion generation have emerged.
We evaluate our method against a comprehensive set of state-of-the-art motion generation models across three major categories: autoregressive models, diffusion-based models, and hybrid architectures. These baselines span both task-specific and multi-task systems, enabling a fair comparison across text-to-motion generation, motion editing, and interleaved instruction following.

Autoregressive baselines include T2M-GPT~\cite{zhang2023t2mgptgeneratinghumanmotion} and the MotionGPT series, which synthesize motion as discrete token sequences in an autoregressive manner. While these models demonstrate strong performance on text-to-motion generation, they typically rely on fixed conditioning formats, making them less suited for handling free-form interleaved instructions. We additionally include MotionLLM~\cite{wu2024motionagentconversationalframeworkhuman} and MG-MotionLLM~\cite{wu2025mgmotionllmunifiedframeworkmotion}, which utilize language backbones to perform unified motion understanding and generation. However, these models primarily operate under simplified contexts and struggle to generalize to complex multi-source motion conditioning.

Diffusion-based baselines, including MLD~\cite{chen2023executingcommandsmotiondiffusion}, SALAD~\cite{hong2025saladskeletonawarelatentdiffusion}, MotionReFit~\cite{jiang2025dynamic}, and MDM~\cite{tevet2022humanmotiondiffusionmodel}, generate high-quality and physically consistent motions through iterative refinement in the latent space. Despite their generative fidelity, they lack instruction-level flexibility as they depend on task-specific conditioning structures, preventing seamless adaptation to diverse generation goals without modifying model architectures.

Hybrid models, such as MotionGPT3~\cite{zhu2025motiongpt3humanmotionsecond}, combine autoregressive sequence modeling with diffusion or contrastive objectives to improve semantic consistency and editing controllability. While these models offer stronger cross-task generalization than single-paradigm systems, they still assume structured conditioning signals and fall short when confronted with unstructured, interleaved motion–text inputs.

\subsection{Benchmarks}

We evaluate OmniMoGen on two representative benchmarks, HumanML3D~\cite{guo2022generating} and MotionFix~\cite{athanasiou2024motionfixtextdriven3dhuman}, each targeting different aspects of motion generation. These benchmarks allow us to assess text-conditioned synthesis, motion refinement under language constraints, and cross-task generalization beyond isolated generation settings.

HumanML3D is a large-scale benchmark for text-to-motion generation that pairs 3D human motion sequences with natural language descriptions. The dataset contains approximately 14K motion clips and nearly 45K textual descriptions, where each motion is annotated with 3–4 sentences collected from human annotators. All motions are represented using a unified 263-dimensional joint format derived from AMASS motion capture sequences, downsampled to 20 fps, and spanning diverse human activities including locomotion, sports, and daily interactions. The evaluation protocol measures how well a model generates motions aligned with free-form textual descriptions rather than structured task prompts. Since OmniMoGen naturally treats single-turn text instructions as interleaved input without requiring specialized modules, HumanML3D serves as a direct test of language grounding and semantic synthesis in the absence of motion context.

MotionFix is designed for text-driven motion editing, where the challenge lies not in generating motion from scratch but in modifying an existing motion sequence based on a natural language instruction. Each instance consists of a source motion, a target motion, and an edit description specifying how the original motion should be altered, such as adjusting movement direction, modifying style, or inserting a new motion segment at a specific temporal interval. Motions are stored in SMPL parameter space, requiring conversion when evaluated under the HumanML3D motion representation. To assess editing quality, MotionFix evaluates both the preservation of unedited attributes and the accuracy of the applied changes by computing Edited-to-Source and Edited-to-Target retrieval metrics. This benchmark emphasizes controlled refinement rather than unconditional generation, making it suitable for validating whether OmniMoGen can selectively modify motion through free-form editing instructions.

\section{Prompts}

\subsection{Prompt for Motion Segment Captioning}

\begin{tcolorbox}[
  colback=gray!5,
  colframe=blue!60!black,
  title=Example Prompt,
  listing only,
  breakable
]
\begin{minted}[
  fontsize=\footnotesize,
  breaklines=true,
  breakanywhere=true,
  breaksymbolleft={},
  breaksymbolright={}
]{yaml}
system: |-
  You are an expert motion analyst.
  Your task is to describe human motion based solely on a provided base64-encoded video.
  Focus strictly on observable motion features without inference or hallucination.

user: |-
  ### Instruction
  Describe the motion shown in the video using a concise natural-language caption.
  Focus on:
  - Main action performed
  - Movement style (if applicable)
  - Spatial trajectory or direction changes
  - Notable transitions or key poses

  ### Constraints
  - Do not guess motivations, environment, or identity.
  - Describe only what is visually observable.

  ### Output Format
  {caption}

  ### Input Video
  {video_base64}
\end{minted}
\end{tcolorbox}

\subsection{Prompt for Annotating Reflection CoT}

\begin{tcolorbox}[
  colback=gray!5,
  colframe=blue!60!black,
  title=Example Prompt,
  listing only,
  breakable
]
\begin{minted}[
  fontsize=\footnotesize,
  breaklines=true,
  breakanywhere=true,
  breaksymbolleft={},
  breaksymbolright={}
]{yaml}
system: |-
  You are an expert motion evaluator.
  Your task is to analyze whether a motion video matches a given natural-language description.
  Provide a detailed step-by-step reasoning process. 
  Only describe what is visually observable from the motion.

user: |-
  ### Task
  Analyze the motion and produce a structured chain of thought including:
  1. A concise description of the motion observed in the video.
  2. A comparison between the observed motion and the provided description.
  3. Evidence-backed statements indicating why they match or do not match.

  ### Constraints
  - Do not infer identity, environment, or intentions.

  ### Output Format
  {cot}

  ### Input Video
  {video_base64}

  ### Target Description
  {caption}
\end{minted}
\end{tcolorbox}

\subsection{Prompt for Annotating Fine-Grained Action Information}

\begin{tcolorbox}[
  colback=gray!5,
  colframe=blue!60!black,
  title=Example Prompt,
  listing only,
  breakable
]
\begin{minted}[
  fontsize=\footnotesize,
  breaklines=true,
  breakanywhere=true,
  breaksymbolleft={},
  breaksymbolright={}
]{yaml}
system: |-
  You are an expert motion annotation system.
  Your task is to analyze a base64-encoded human motion clip and output a structured action list.

user: |-
  ### Task
  Extract a list of fine-grained actions from the video. 
  Each action must include the following fields:
  - action_type
  - body_part
  - style
  - duration
  - trajectory

  ### Constraints
  - Output strictly based on visible motion only.
  - Use only the predefined category options.
  - If uncertain, return "unknown".
  - Do not infer environment, identity, or camera details.

  ### Output Format (JSON Array)
  [
    {
      "action_type": "{action_type}",
      "body_part": "{body_part}",
      "style": "{style}",
      "duration": "{duration}",
      "trajectory": "{trajectory}"
    }
  ]

  ### Predefined Options
  action_type: {action_type_options}
  body_part: {body_part_options}
  style: {style_options}
  trajectory: {trajectory_options}

  ### Input Video
  {video_base64}
\end{minted}
\end{tcolorbox}

\end{document}